\documentclass[11pt, letterpaper]{article}

\usepackage[left=1in, right=1in, top=1in, bottom=1in]{geometry}
\usepackage[utf8]{inputenc}
\usepackage[T1]{fontenc}
\usepackage{bm}
\usepackage{type1cm}
\usepackage{lettrine}
\usepackage{amsmath,amssymb,amsthm}
\usepackage{moreverb}
\usepackage{mathtools}
\usepackage{amsmath}
\usepackage{amssymb}
\usepackage{algorithmic}
\usepackage{graphics}
\usepackage{graphicx}
\usepackage{subcaption}
\usepackage{caption}
\usepackage{extarrows}
\usepackage{color}
\usepackage{framed}
\usepackage{wrapfig}
\usepackage{bm}
\usepackage{mathrsfs}
\usepackage{mathabx}
\usepackage{multirow}
\usepackage{longtable}
\usepackage{hyperref}
\usepackage{paralist}
\usepackage{indentfirst}
\usepackage{relsize}
\usepackage{extarrows}
\usepackage{upgreek}
\usepackage{bm}
\usepackage{mwe}
\usepackage[dvipsnames,table]{xcolor}
\usepackage{booktabs}
\usepackage{authblk}
\usepackage{lettrine}
\usepackage{type1cm}
\usepackage{threeparttable}
\usepackage[sort&compress,numbers]{natbib}
\usepackage[figurename=Figure]{caption}
\usepackage[normalem]{ulem}
\usepackage{adjustbox}
\usepackage{ulem}
\usepackage{xcolor}
\usepackage[most]{tcolorbox}

\newtcolorbox{verbatimbox}[1][]{
  colback=brown!10!white, 
  colframe=gray!75!black, 
  arc=2mm, 
  boxrule=0.5pt, 
  title=#1, 
  fonttitle=\bfseries,
  breakable, 
  enhanced, 
  top=2mm, bottom=2mm, left=2mm, right=2mm, 
  fontupper=\ttfamily 
}

\graphicspath{ {./Figure/} }
\usepackage[font=footnotesize,labelfont=bf]{caption}

\newcommand{\graph}[1]{\mathcal{}}
\providecommand{\keywords}[1]{\textbf{\textit{Keywords: }} #1}

\hypersetup{
bookmarks=true,
bookmarksopen=true,
bookmarksnumbered=true,
unicode=false,
pdftoolbar=true,
pdfmenubar=true,
pdffitwindow=false,
pdfstartview={FitH},
pdftitle={My title},
pdfauthor={Author},
pdfsubject={Subject},
pdfcreator={Creator},
pdfproducer={Producer},
pdfkeywords={keywords},
pdfnewwindow=true,
colorlinks=true,
linkcolor=blue,
citecolor=blue,
filecolor=blue,
urlcolor=blue
}

\begin{document}

\title{\textbf{A closed-loop framework for perovskite precursor additive discovery}}
\author[1,3,$\dag$]{Xin-De Wang}
\author[2,$\dag$] {Zhi-Rui Chen}
\author[1,3,$\dag$*]{Ze-Feng Gao}
\author[1,3]{Peng-Jie Guo}
\author[2,*] {Cheng Mu}
\author[1,3,*]{Zhong-Yi Lu}

\affil[1]{\small School of Physics and Key Laboratory of Quantum State Construction and Manipulation (Ministry of Education), Renmin University of China, Beijing 100872, China}
\affil[2]{\small School of Chemistry and Life Resource, Renmin University of China, Beijing, China }
\affil[3]{\small AI-Driven Quantum Materials Research Center, School of Physics, Renmin University of China\vspace{18pt}}

\affil[*]{e-mail: zfgao@ruc.edu.cn; cmu@ruc.edu.cn; zlu@ruc.edu.cn}

\affil[$\dag$]{These authors contributed equally}

\date{}

\maketitle

\normalsize

\vspace{-28pt} 

\begin{abstract}
\small
Efficient discovery of precursor additives is essential for improving the performance of perovskite solar cells, yet the large chemical space makes conventional trial-and-error screening inefficient.
{Here we show that LEAP~(LLM-driven Exploration via Active Learning for Perovskites), a closed-loop framework combining a domain-specialized language model with active learning, prioritizes perovskite precursor additives under low-data conditions. The model extracts mechanism-relevant knowledge from the literature and converts it into interpretable molecular descriptors for uncertainty-aware Bayesian optimization. On an evaluation drawn from unseen literature, the specialized model provides more mechanism-consistent answers than the tested general-purpose models. In an expert-guided proof-of-concept study, two later-round additives increase average device power conversion efficiencies to 20.13\% and 20.87\%, compared with 19.25\% for the control, and the best device reaches 21.32\%. Because only a limited number of candidates are tested prospectively, these results provide preliminary evidence that literature-grounded molecular representations, Bayesian optimization, and expert assessment can support additive discovery for perovskite solar cells.}
\end{abstract}

\vspace{12pt} 

\section*{Introduction}\label{sec1}
Perovskite solar cells~(PSCs) have rapidly emerged as one of the most promising photovoltaic technologies because of their outstanding optoelectronic properties, solution processability, and rapidly rising power conversion efficiencies~(PCEs)~\cite{kojima2009organometal,gratzel2014light,jiang2024rapid}. However, despite this progress, the practical development of PSCs remains constrained by the intrinsic complexity of perovskite crystallization and defect formation~\cite{liu2023stabilization}. During film formation, undercoordinated ions, halide vacancies, and interfacial trap states are readily generated in the bulk and at surfaces~\cite{xu2025point}, which exacerbate non-radiative recombination, reduce operational stability, and ultimately limit device performance~\cite{xiong2025homogenized,jiang2023towards,tao2025suppressing,zhu2023longterm}.

Additive engineering has become one of the most effective strategies to address these challenges~\cite{zhang2023tailoring}. By incorporating functional molecules into precursor solutions, it is possible to regulate crystallization kinetics, modulate intermediate phases~\cite{zhou2026additive}, passivate bulk and interfacial defects~\cite{teale2024molecular}, and improve film morphology and energy-level alignment~\cite{wang2026homogenized,isikgor2023molecular}. As a result, additive design has played a central role in advancing both the efficiency and stability of PSCs. Yet the chemical space of potentially useful additives is extremely broad, spanning diverse functional groups, interaction motifs, and physicochemical properties. This combinatorial complexity makes experimental exploration highly resource-intensive, while conventional trial-and-error screening is poorly matched to the scale of the search problem~\cite{cheng2020evolutionary,herbol2018efficient}.

Recent advances in artificial intelligence~(AI), including AI-driven inverse design frameworks, offer a potential route to more efficient materials discovery~\cite{chen2026large,jablonka2024leveraging,Han2025AIDrivenInverseDesign}. \textcolor{black}{In PSC research, this trend has progressed from descriptor-based machine-learning~(ML) screening to deep-learning~(DL) representation learning and, more recently, LLM-assisted literature mining and screening. For example, descriptor-based ML has been used to relate molecular features of ammonium salts to interface-passivation-induced PCE improvements and to screen candidate passivation materials from chemical databases~\cite{Zhi2023MLPassivation}, while DL frameworks have enabled automated extraction of morphology descriptors from perovskite SEM images for device-performance prediction~\cite{Zhou2025SelfUNet}. Recent LLM-assisted workflows further show that fragmented perovskite literature can be converted into structured datasets and combined with uncertainty-aware predictive models for chemical-space exploration~\cite{Marshall2026POLARIS}.} However, existing approaches remain limited for additive development in PSCs~\cite{Wang2026PerovskiteAdditiveAI}. Most machine-learning frameworks in perovskite research are designed for property prediction within predefined datasets~\cite{tao2021machine}, while general-purpose large language models~(LLMs) lack sufficiently specialized mechanistic understanding of additive-precursor interactions, defect chemistry, and crystallization control~\cite{wang2026perovskite}. What is needed is not only predictive ranking, but a domain-informed system that can translate literature-scale chemical knowledge into experimentally actionable hypotheses and iteratively improve from real validation outcomes~\cite{dagdelen2024structured}.

Here we develop LEAP~(LLM-driven Exploration via Active Learning for Perovskites), an iterative framework for perovskite precursor additive discovery that couples a domain-specialized LLM, Perovskite-RL, with active learning~(AL)~\cite{kusne2020onthefly}. In LEAP, Perovskite-RL is trained to extract mechanism-relevant knowledge from the additive literature and to represent candidate molecules through interpretable descriptors linked to additive function in PSCs. These knowledge-guided descriptors are then integrated into a Bayesian Optimization~(BO) workflow to prioritize candidates for experimental validation, while newly obtained experimental results are fed back into the model loop to refine subsequent selection. In this way, LEAP implements an expert-in-the-loop closed workflow that combines domain-specific reasoning, uncertainty-aware prioritization, experimental feasibility review, and feedback from device validation. In this proof-of-concept study, the workflow prioritized additives across three screening rounds and identified later-round candidates that improved device-level performance relative to the control. Rather than establishing a fully autonomous or generally optimized discovery platform, these results show how literature-grounded mechanistic representations can be coupled with Bayesian optimization to support experimentally actionable additive prioritization under low-data conditions.

Distinct from using LLMs only for text mining or molecular description, LEAP follows the broader direction of language-representation-based materials discovery by converting literature-derived mechanistic reasoning into interpretable descriptors that can be coupled to Bayesian optimization and experimental feedback~\cite{qu2024leveraging}.

\section*{Results}\label{sec-results}
\subsection*{Framework of LEAP}
\begin{figure*}[t]
    \centering
    \includegraphics[width=\textwidth]{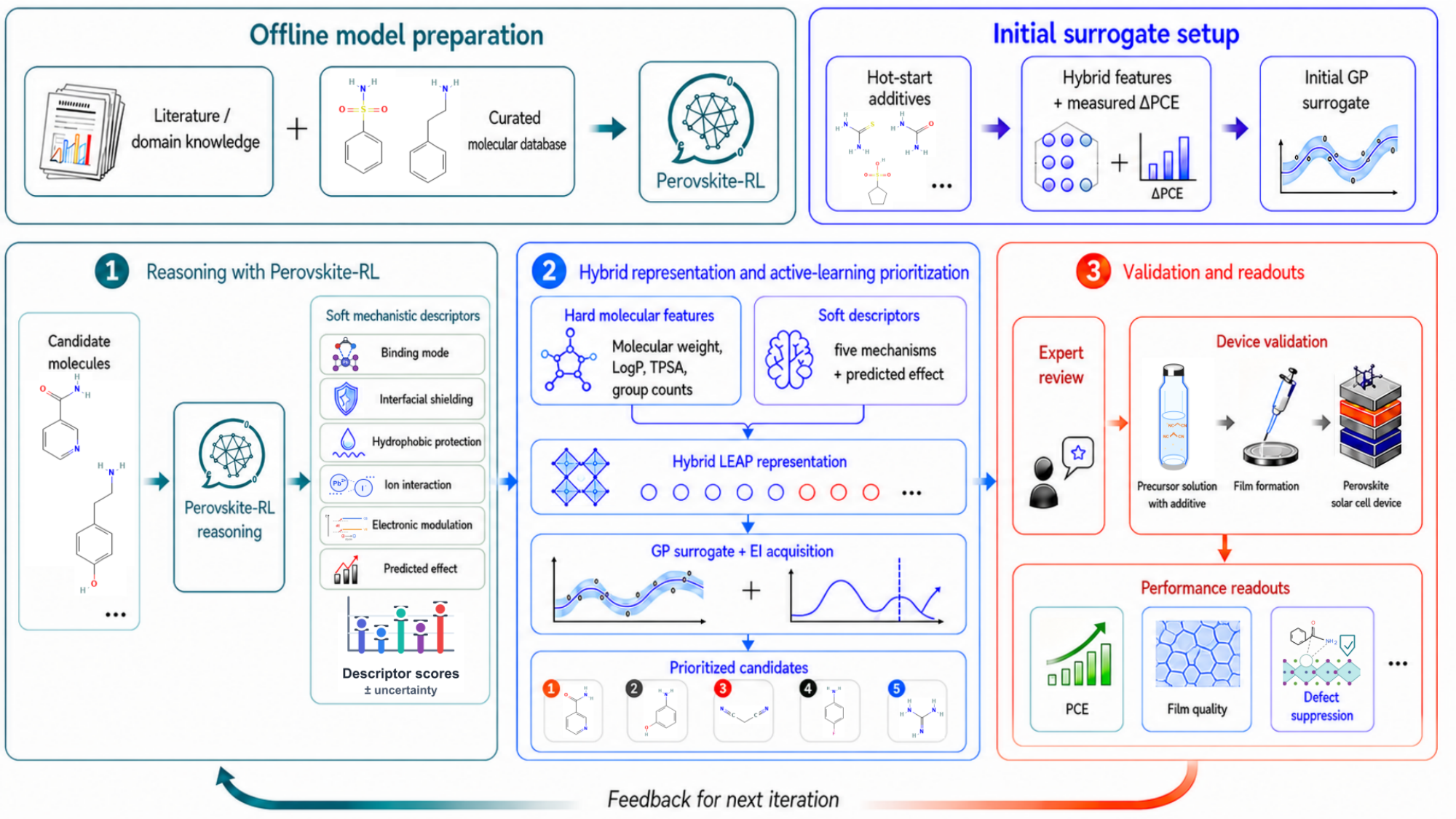}
    \caption{\textcolor{black}{Overview of the LEAP expert-in-the-loop workflow for perovskite precursor additive prioritization, with arrows indicating the information flow across model construction, candidate scoring, expert review, experimental validation, and feedback updating.} Literature-derived domain knowledge and a curated molecular database are used to construct Perovskite-RL, while hot-start additives with measured \(\Delta\mathrm{PCE}\) values initialize the GP surrogate model. In each iteration, Perovskite-RL evaluates candidate molecules and generates soft mechanistic descriptors, which are integrated with hard molecular features to form a hybrid LEAP representation for GP-based expected-improvement acquisition. Prioritized candidates are reviewed by domain experts and experimentally validated in perovskite solar cells. Device-level readouts, including PCE, film quality, and defect suppression, are fed back to update the dataset, descriptors, and surrogate model for the next iteration.}
    \label{fig:fig-1}
\end{figure*}

LEAP is an expert-in-the-loop, closed-workflow framework for perovskite precursor additive prioritization that integrates literature-grounded mechanistic reasoning with uncertainty-aware experimental selection~\cite{hase2019nextgeneration}. In this proof-of-concept implementation, domain knowledge extracted from the perovskite additive literature is combined with active learning to prioritize candidate molecules~\cite{lookman2019active}, guide experimental validation, and update subsequent rounds of selection. 
\textcolor{black}{The overall workflow is summarized in Figure~\ref{fig:fig-1}}, 
\textcolor{black}{which follows the information flow of each LEAP round from literature and candidate-library input, to Perovskite-RL mechanistic evaluation, hybrid descriptor construction, GP/EI-based prioritization, expert feasibility review, experimental validation, and feedback-driven updating.}

At the core of LEAP is a domain-specialized LLM, Perovskite-RL, which is trained through supervised fine-tuning~(SFT) and reinforcement learning~(RL) on the perovskite precursor additive literature~\cite{choi2024accelerating}. Perovskite-RL is used to assess candidate molecules in a mechanism-relevant manner and to generate descriptors linked to additive function. These descriptors are then combined with molecular features and used as the input to a Bayesian optimization~(BO) workflow~\cite{cisse2026automate}. Specifically, a Gaussian process~(GP) regressor serves as the probabilistic surrogate model~\cite{rasmussen2006gaussian} to estimate both the expected performance gain and the associated predictive uncertainty for untested candidates, and the expected improvement~(EI) acquisition function is used to prioritize molecules for each round of selection~\cite{jones1998efficient}. 
At the current stage, LEAP operates in an expert-in-the-loop mode: model-prioritized candidates are not automatically sent for validation, but are first reviewed by domain experts for chemical plausibility, commercial availability, precursor-solution compatibility, safety, cost, and device-fabrication feasibility.
Newly obtained experimental results are then fed back into the workflow to refine the next round of prioritization. In each feedback cycle, the newly measured device performance is appended to the existing hot-start dataset and used together with previous experimental data to retrain the GP surrogate model. At the same time, the prompting template used for Perovskite-RL evaluation is updated to incorporate the latest experimental observations, so that soft mechanistic descriptors for the accumulated training set and candidate pool are regenerated under the updated reasoning context. The retrained surrogate and updated descriptors are then used to recompute acquisition scores for the next round of candidate selection.

Because LEAP relies on literature-grounded mechanistic assessments as the basis for downstream candidate representation and prioritization, the reliability of this reasoning component must first be established. 
\textcolor{black}{We therefore constructed a 32-question mechanism-consistency evaluation set using 11 recently selected studies} on perovskite precursor additives that were strictly excluded from the training corpus. From these unseen papers, we curated 32 multiple-choice questions to test whether the model could generate mechanistically consistent judgments aligned with the interpretations reported in the original literature.

\begin{figure}[htbp]
    \centering
    \includegraphics[width=1.0\textwidth]{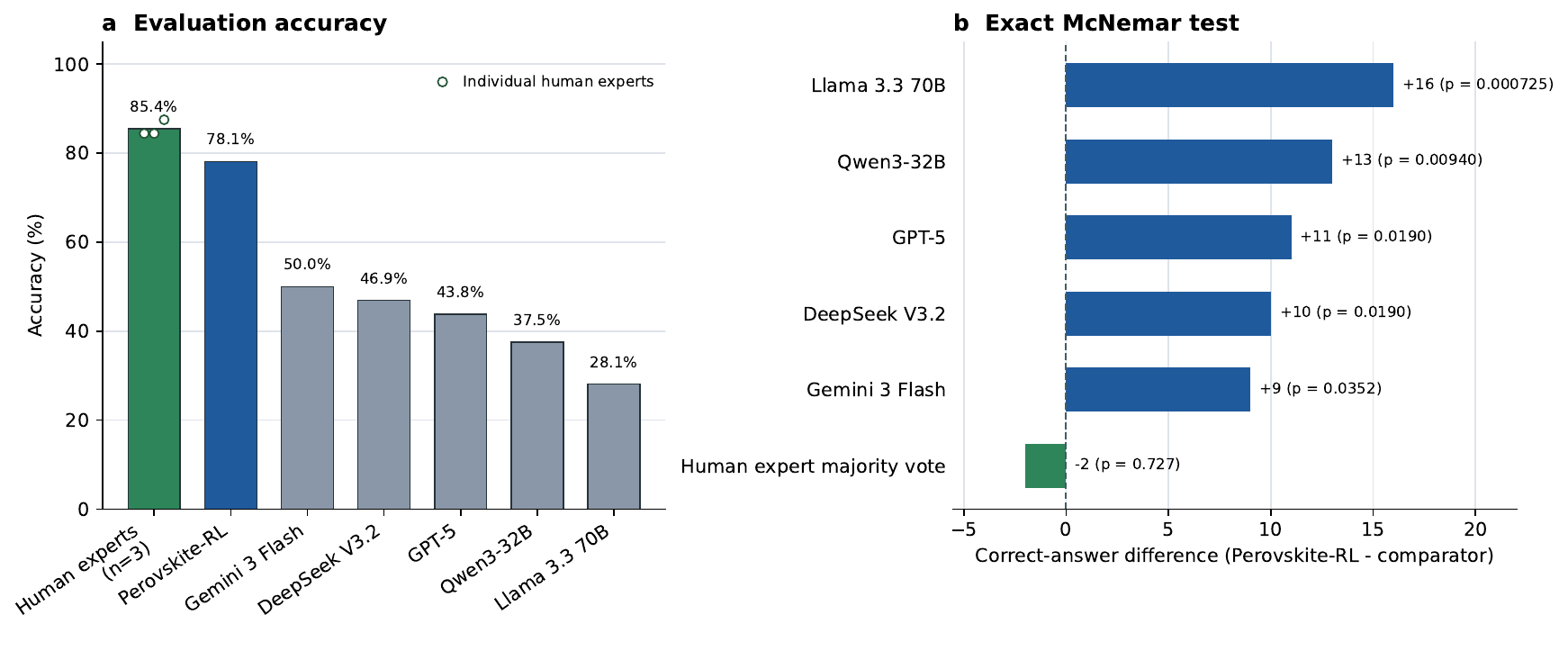}
    \caption{\textcolor{black}{Mechanism-consistency evaluation with model and human-expert references. (a) Observed accuracy of Perovskite-RL, three human experts, and baseline models on 32 questions from 11 unseen additive papers, with bars ordered by observed accuracy. The human-expert bar shows the mean accuracy across three evaluators, and the three white circles on this bar show the accuracies of the three individual human experts. (b) Pairwise two-sided exact McNemar comparisons between Perovskite-RL and each model baseline, together with a paired comparison against the human-expert majority-vote reference. Positive values indicate a net correct-answer advantage for Perovskite-RL, whereas negative values indicate a net advantage for the comparator. Model-baseline P values were Holm-Bonferroni adjusted across the five model comparisons; the human-expert majority-vote comparison is reported as a reference comparison.}}
    \label{fig:fig-2}
\end{figure}

As shown in Figure~\ref{fig:fig-2}a, 
the reasoning component used in LEAP answered 25 of 32 questions correctly, corresponding to an accuracy of 78.1\%, with a 95\% Wilson confidence interval from 61.2\% to 89.0\%. \textcolor{black}{As a human-expert reference, three human experts independently answered the same 32 questions under a closed-book setting, achieving 27/32, 27/32, and 28/32 correct answers, respectively.} This accuracy was higher than those of the evaluated backbone and general-purpose baseline models \textcolor{black}{on the same mechanism-consistency evaluation set.}
\textcolor{black}{Thus, Perovskite-RL approached but did not exceed the human-expert reference, while outperforming the evaluated model baselines.} Pairwise exact McNemar tests were used to compare Perovskite-RL with each \textcolor{black}{model} baseline, followed by Holm-Bonferroni correction across the five pairwise comparisons, as shown in Figure~\ref{fig:fig-2}b~\cite{mcnemar1947note}. \textcolor{black}{A separate paired comparison against the human-expert majority vote yielded a net difference of -2 correct answers for Perovskite-RL and was not significant~(P = 0.73).} Because \textcolor{black}{this question set} contains only 32 paired questions, these statistical results should be interpreted as a focused sanity check of mechanism-consistent reasoning rather than as an exhaustive model ranking. This \textcolor{black}{evaluation} therefore provides preliminary support for using Perovskite-RL as the reasoning module for downstream candidate representation and prioritization.

\subsection*{Mechanistic representation in LEAP}
Conventional molecular descriptors can capture basic physicochemical properties of candidate additives~\cite{moriwaki2018mordred}, but they do not explicitly encode the higher-level mechanisms through which additives influence perovskite crystallization, defect passivation, interfacial protection, and environmental stability~\cite{vasilopoulou2020molecular}. For additive discovery in PSCs, however, these mechanism-level effects are often more directly related to function than structural descriptors alone. LEAP therefore constructs a mechanism-aware representation of candidate molecules that is designed for prioritization in a closed-loop discovery setting rather than for structural description alone.

To achieve this, LEAP derives soft mechanistic descriptors from literature-grounded reasoning over candidate molecules~\cite{gupta2022matscibert}. Instead of directly using high-dimensional latent embeddings from the LLM, which are difficult to interpret and may obscure the contribution of physically meaningful features~\cite{gallegos2024explainable}, LEAP represents each molecule through explicit scores along five mechanism-relevant dimensions: binding mode, interfacial shielding, hydrophobic protection, ion-interaction potential, and electronic modulation~\cite{deng2020coordination,thiesbrummel2026ion}. In this way, domain knowledge extracted from the additive literature is translated into an interpretable descriptor space that is directly linked to additive function in PSCs.

To improve robustness, these descriptors are obtained through repeated probabilistic reasoning and aggregation~\cite{wang2023selfconsistency}, allowing the representation to capture not only dominant mechanistic tendencies but also inference variability across candidates. The resulting soft descriptors are then integrated with deterministic hard features, including standard molecular physicochemical properties, to form a hybrid representation for downstream modelling. By combining explicit molecular constraints with mechanism-level semantic information, LEAP establishes a feature space suitable for subsequent uncertainty-aware candidate prioritization.

\subsection*{LEAP for candidate prioritization}

To identify experimentally actionable additives under low-data conditions, LEAP uses the hybrid molecular representation as the input space for surrogate modelling~\cite{wang2022bayesian}. Starting from a hot-start set of experimentally characterized additives, a GP surrogate model is trained to map the combined soft and hard descriptors to the target performance metric. This probabilistic surrogate allows LEAP to estimate, for each untested candidate, both the expected performance gain and the associated predictive uncertainty~\cite{zhang2022uncertainty}. In this way, candidate evaluation is performed not only in terms of predicted benefit, but also in terms of how confidently that benefit can be assessed from the currently available data.

Candidate prioritization is then guided by the EI acquisition function, which balances exploitation of molecules predicted to perform well against exploration of under-sampled regions of the candidate space. LEAP therefore does not simply rank candidates by predicted mean response, but instead prioritizes molecules that are jointly promising and informative for subsequent rounds of optimization. At the current stage, this process operates in an expert-in-the-loop mode, that is, model-prioritized candidates are further reviewed for chemical plausibility and experimental feasibility before validation~\cite{adams2024human}. By combining uncertainty-aware model guidance with expert assessment, LEAP converts mechanism-aware molecular representation into an experimentally practical strategy for iterative additive prioritization. \textcolor{black}{Thus, the prospective validation is intended to evaluate the practical expert-in-the-loop workflow, rather than a fully autonomous model-only selection process.}

\begin{figure}[t]
    \centering
    \includegraphics[width=\textwidth]{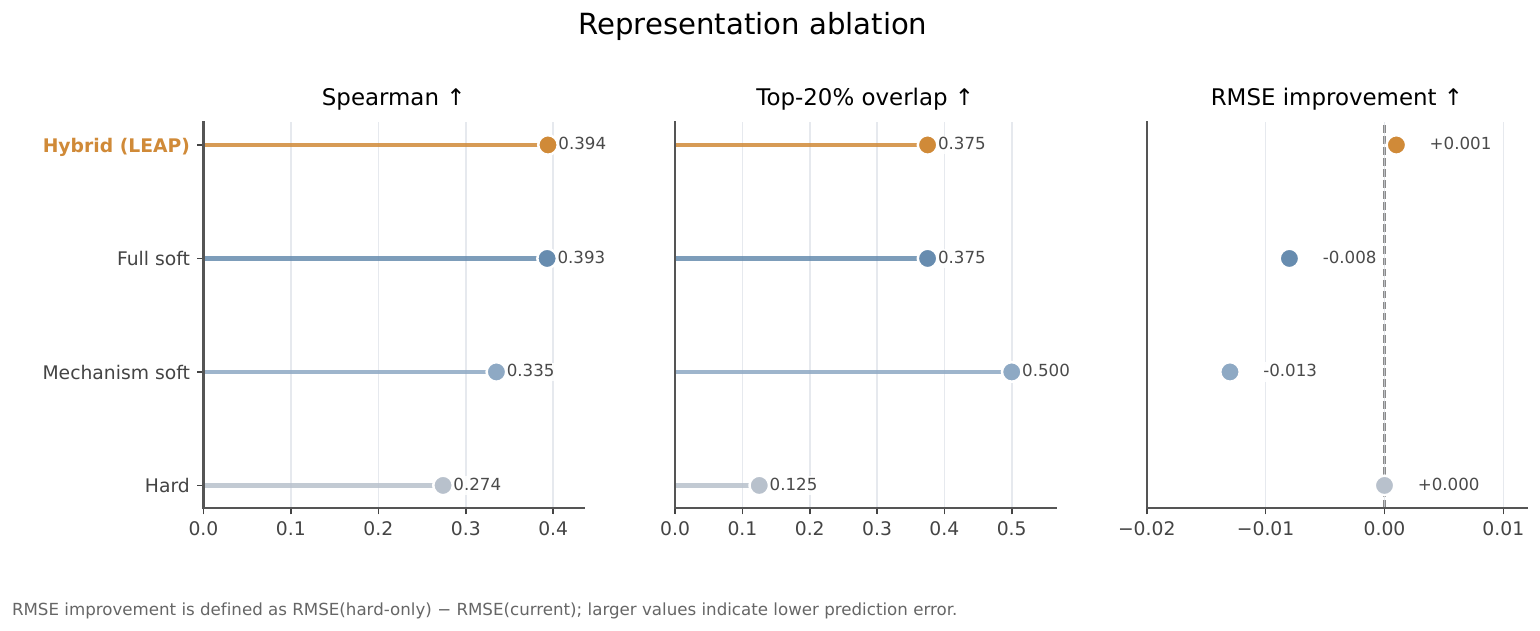}
    \caption{Retrospective representation ablation for LEAP candidate prioritization. GP surrogate models were evaluated on 36 hot-start additives using relative PCE change as the target. Hard features, mechanism-aware soft descriptors, full soft descriptors, and the hybrid LEAP representation were compared using Spearman correlation, top-20\% overlap, and RMSE improvement relative to the hard-feature baseline. Larger RMSE improvement indicates lower prediction error. \textcolor{black}{The four representation settings correspond to hard molecular descriptors, mechanism-level soft descriptors, full soft descriptors including predicted effect, and the combined LEAP hybrid representation.}
    }
    \label{fig:fig-3}
\end{figure}

To examine whether the representation design contributes useful information beyond conventional molecular descriptors~\cite{rajabikochi2025adaptive}, we performed a retrospective representation ablation on the 36 experimentally characterized hot-start additives. We compared GP surrogate models trained with conventional hard molecular features~\cite{liang2021benchmarking}, mechanism-aware soft descriptors derived from repeated probabilistic reasoning, full soft descriptors that additionally include the overall predicted-effect score, and the hybrid LEAP representation combining hard and soft information. 
\textcolor{black}{Here, hard molecular features refer to deterministic structure-derived descriptors such as physicochemical properties and functional-group counts; mechanism-aware soft descriptors refer to the five Perovskite-RL-derived scores for binding mode, interfacial shielding, hydrophobic protection, ion-interaction potential, and electronic modulation; full soft descriptors further include the overall predicted-effect score; and the hybrid LEAP representation concatenates the hard and full soft descriptors.}
Model performance was evaluated using Spearman rank correlation, top-20\% overlap, and RMSE improvement relative to the hard-feature baseline, thereby capturing ranking quality, recovery of high-performing candidates, and prediction error, respectively, as shown in Figure~\ref{fig:fig-3}~\cite{rohr2020benchmarking}. Bootstrap confidence intervals for these metrics are provided in Table~S3.

The representation ablation suggests that literature-derived soft descriptors provide prioritization-relevant information beyond hard molecular features alone, although the effect size is modest and metric-dependent. In the 36-additive retrospective analysis, the hard-feature baseline gave a Spearman correlation of 0.274 and a top-20\% overlap of 0.125, whereas the hybrid LEAP representation gave a Spearman correlation of 0.394 and a top-20\% overlap of 0.375. These results support the possible value of mechanism-aware information for candidate prioritization under low-data conditions~\cite{xu2023small}. The hybrid representation also showed only a small RMSE improvement relative to the hard-feature baseline. We therefore use the hybrid representation as a balanced and chemically interpretable feature space, but do not interpret it as evidence of uniform superiority across all metrics.

\subsection*{Experimental validation of LEAP-selected additives}

\begin{figure*}[t]
    \centering
    \includegraphics[width=\textwidth]{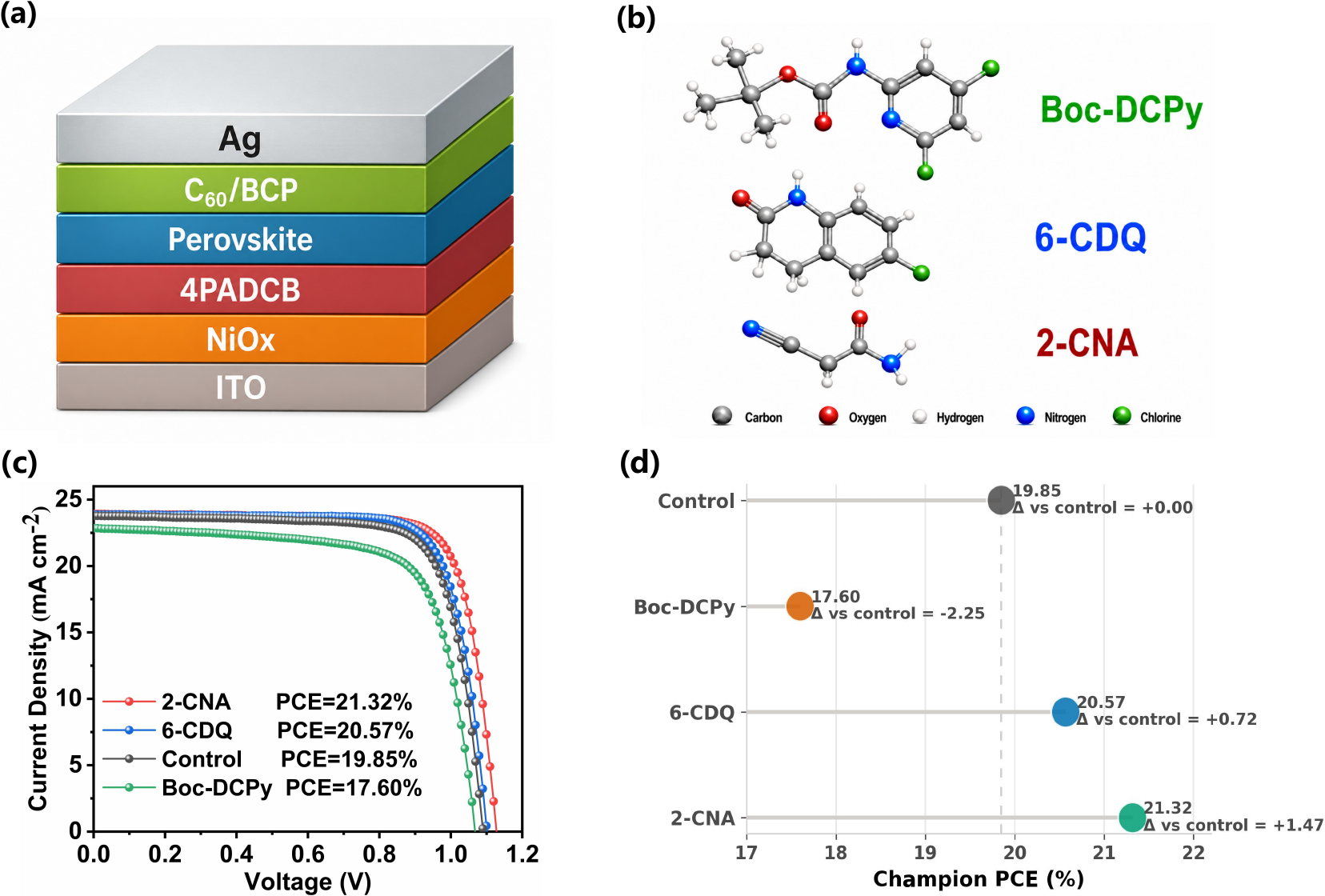}
    \caption{LEAP-prioritized additives and representative champion-device results. (a) Device architecture. (b) Molecular structures of Boc-DCPy, 6-CDQ, and 2-CNA. (c) Representative \textit{J}-\textit{V} curves. (d) Champion PCE values for the control and the three expert-reviewed LEAP validation rounds.
    }
    \label{fig:fig-4}
\end{figure*}

To examine the practical behavior of LEAP-guided prioritization in an expert-in-the-loop validation setting, we selected one representative candidate from each round for device testing. Figure~\ref{fig:fig-4}a,b show the inverted device architecture and the molecular structures of the tested LEAP-prioritized additives. Across three consecutive rounds, Boc-DCPy, 6-CDQ, and 2-CNA were selected after model-guided prioritization and expert feasibility review for testing in inverted perovskite solar cells with the structure ITO/NiO$_x$/4PADCB/perovskite/C$_{60}$/BCP/Ag~\cite{li2026efficient}. Figure~\ref{fig:fig-4}c shows the representative \textit{J}–\textit{V} responses obtained from the three validation rounds. The first-round candidate, Boc-DCPy, did not improve device efficiency and yielded a champion PCE of 17.60\%, lower than the 19.85\% champion PCE of the control device. After the Boc-DCPy result was incorporated into the accumulated training set, the surrogate model was retrained and the Perovskite-RL prompting template was updated to reflect the newly observed experimental outcome. In the second round, the updated workflow prioritized 6-CDQ, which yielded a champion PCE of 20.57\%. In the third round, 2-CNA yielded the highest champion-device performance among the tested additives, with a PCE of 21.32\%, a \textit{V}$_\mathrm{OC}$ of 1.128 V, a \textit{J}$_\mathrm{SC}$ of 23.92 mA cm$^{-2}$, and a fill factor of 0.790. As shown in Figure~S3, the integrated current densities from the external quantum efficiency~(EQE) spectra were consistent with the \textit{J}–\textit{V} results~\cite{christians2015best}. \textcolor{black}{A stabilized power-output measurement of the champion 2-CNA device at 0.84 V for 60 s yielded a stabilized PCE of 19.64\% (Figure~S6).} These representative champion-device results indicate that the later-round candidates were experimentally promising, but population-level device statistics are needed to assess whether the performance trend is reproducible.

\begin{figure}[!t]
    \centering
    \includegraphics[width=\textwidth]{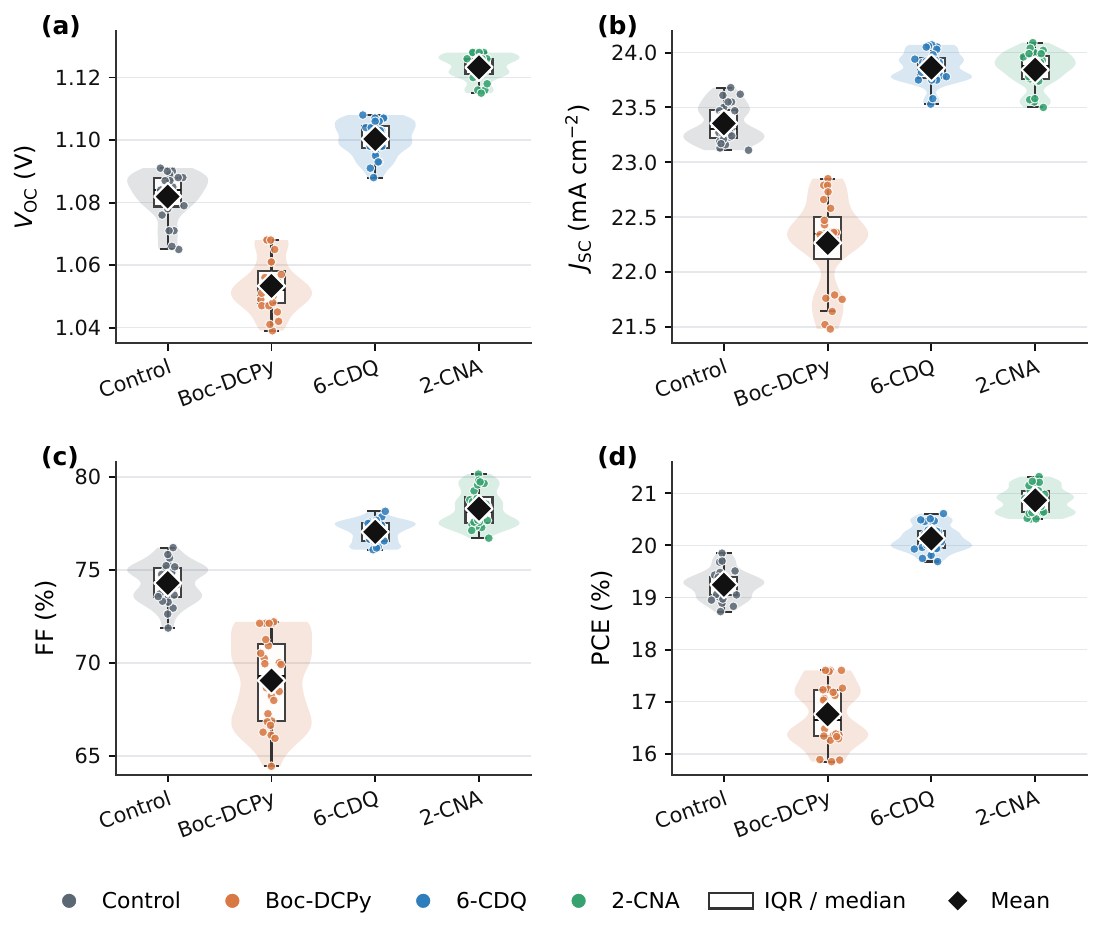}
    \caption{Device statistics for control and additive-treated perovskite solar cells. Distributions of (a) $V_{\mathrm{OC}}$, (b) $J_{\mathrm{SC}}$, (c) FF, and (d) PCE are shown for 24 devices per group\textcolor{black}{, obtained from four independent fabrication batches with six devices per group in each batch}. Colored points denote individual devices, violins show distribution density, boxes indicate interquartile ranges with median lines, and black diamonds indicate mean values.}
    \label{fig:fig-5}
\end{figure}

To assess reproducibility, we fabricated 24 devices for each condition, including the control, Boc-DCPy-, 6-CDQ-, and 2-CNA-treated groups\textcolor{black}{, across four independent fabrication batches conducted on different dates using freshly prepared precursor solutions, with six devices per condition in each batch}. As shown in Figure~\ref{fig:fig-5} and Table~S5, the mean PCE values were 19.25 ± 0.28\% for the control, 16.76 ± 0.58\% for Boc-DCPy, 20.13 ± 0.25\% for 6-CDQ, and 20.87 ± 0.25\% for 2-CNA. Relative to the control, Boc-DCPy decreased the mean PCE by 2.49 percentage points, whereas 6-CDQ and 2-CNA increased the mean PCE by 0.89 and 1.62 percentage points, respectively. Welch’s t-tests using the raw device-level values supported the same trend, with two-sided P values of 2.23 × 10$^{-19}$ for Boc-DCPy, 4.44 × 10$^{-15}$ for 6-CDQ, and 4.78 × 10$^{-25}$ for 2-CNA compared with the control. \textcolor{black}{The direction of the PCE change was reproduced across all four fabrication batches (Figure~S5), providing a batch-level reproducibility check in addition to the pooled device-level comparison.} These population-level results support the performance trend observed in the representative champion devices and provide a more robust basis for evaluating the LEAP-prioritized candidates.

To provide a practical reference for the LEAP-guided validation route, we also evaluated a non-LEAP comparison route. In this route, three representative additives, including 4-hydroxy-8-(trifluoromethyl)quinoline, sodium m-hydroxybenzenesulfonate, and 5-bromo-2-chloro-4-fluoroaniline, were selected by domain experts from the same candidate library used by LEAP. The same availability and experimental-feasibility constraints were applied, and all three additives were evaluated under the same device architecture, fabrication procedure, and testing conditions. The three tested non-LEAP candidates did not show a comparable performance trend and did not reach the device performance obtained for the later-round LEAP-prioritized candidates. Because this comparison involved only three expert-selected candidates and was not a randomized or exhaustive \textcolor{black}{comparison} of the candidate space, it should be interpreted as a practical reference rather than as a definitive comparison between LEAP and expert selection.

\begin{figure}[!t]
    \centering
    \includegraphics[width=\textwidth]{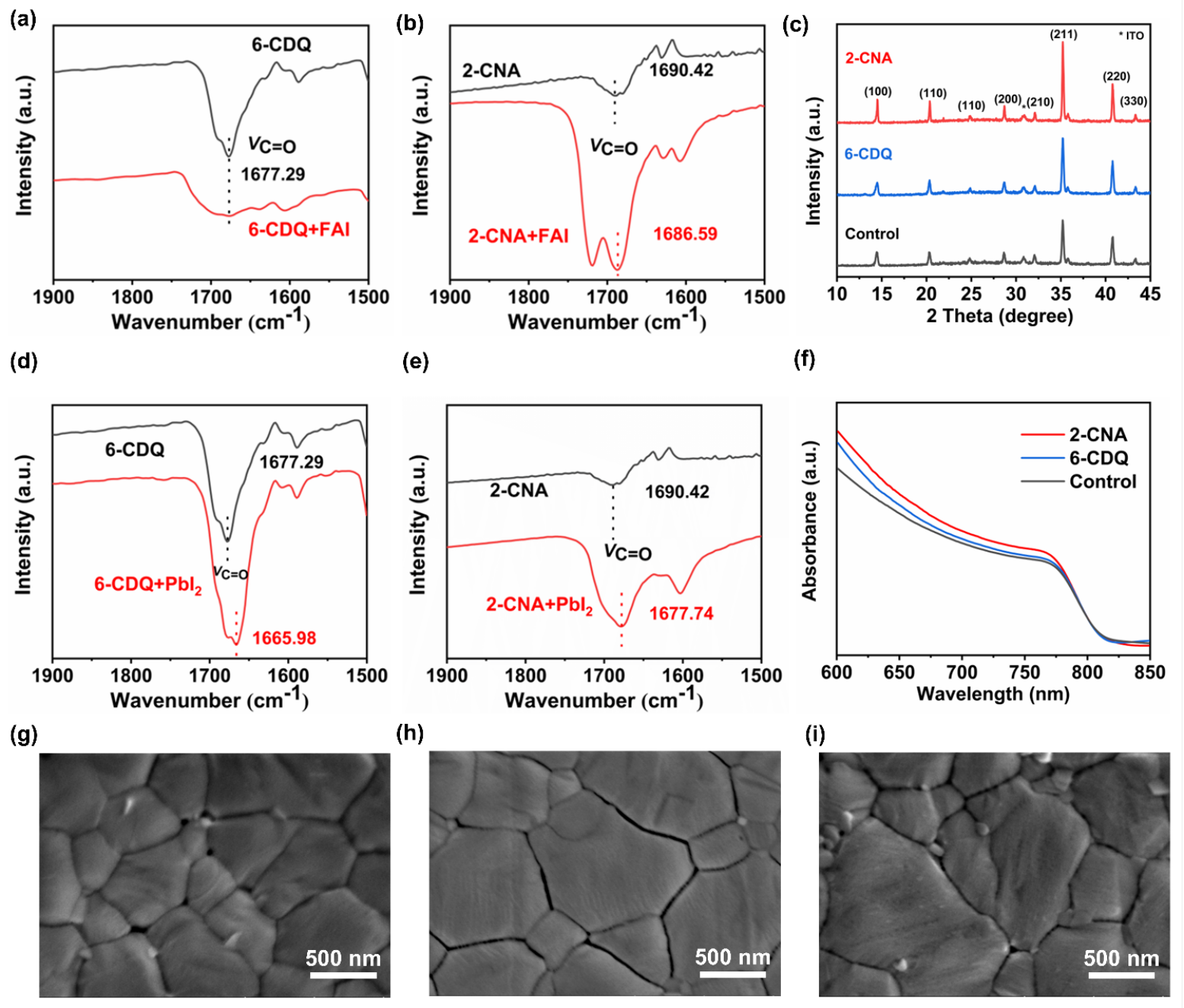}
    \caption{Molecular interactions and film characterization of LEAP-selected additives. (a,b) FTIR spectra of 6-CDQ and 2-CNA with FAI. (c) XRD patterns of control and additive-treated films. (d,e) FTIR spectra of 6-CDQ and 2-CNA with PbI$_2$. (f) UV–Vis absorption spectra. (g–i) Top-view SEM images.
}
    \label{fig:fig-6}
\end{figure}

To examine possible materials-level factors associated with the improved performance of the later-round LEAP-prioritized additives, we next investigated additive-precursor interactions and film properties.
As shown in Figure~\ref{fig:fig-6}a,b,d,e, Fourier-transform infrared (FTIR) measurements showed spectral changes consistent with coordination between both 6-CDQ and 2-CNA and PbI$_2$, whereas 2-CNA also showed spectral evidence of interaction with formamidinium iodide (FAI). This additional interaction is consistent with a broader precursor-interaction profile for 2-CNA~\cite{wang2024manipulating}, which may contribute to more favorable coordination and crystallization behavior. Consistent with this interpretation, Figure~\ref{fig:fig-6}c,f-i show that additive-treated films had enhanced crystallinity, stronger optical absorption, and enlarged grains relative to the control, with the effect most pronounced for 2-CNA. These observations support, but do not by themselves prove, a link between the later-round candidates, precursor interactions, and improved film formation.
\textcolor{black}{Tauc analysis of the UV--Vis absorption data yielded direct optical bandgaps of 1.543, 1.543, and 1.545 eV for the control, 6-CDQ-, and 2-CNA-treated films, respectively (Figure~S9). Thus, the perovskite used here is an approximately 1.55 eV system rather than a wide-bandgap perovskite.}

\textcolor{black}{Additional photophysical and morphological analyses provide complementary evidence for the effects of the additives. The steady-state PL intensity increased from the control to 6-CDQ and 2-CNA, while the TPC decay time decreased from 0.882 $\mu$s for the control to 0.691 $\mu$s for 6-CDQ and 0.596 $\mu$s for 2-CNA, and the TPV lifetime increased from 10.3 to 10.5 and 11.7 $\mu$s, respectively (Figure~S7). SEM-based grain-size analysis gave mean grain sizes of $0.69 \pm 0.21$ $\mu$m for the control, $0.77 \pm 0.30$ $\mu$m for 6-CDQ, and $0.79 \pm 0.26$ $\mu$m for 2-CNA (Figure~S8). These results are consistent with improved film quality, more efficient charge extraction, and reduced non-radiative recombination, but do not independently establish the proposed coordination or passivation mechanism.}

\begin{figure}[t]
    \centering
    \includegraphics[width=\textwidth]{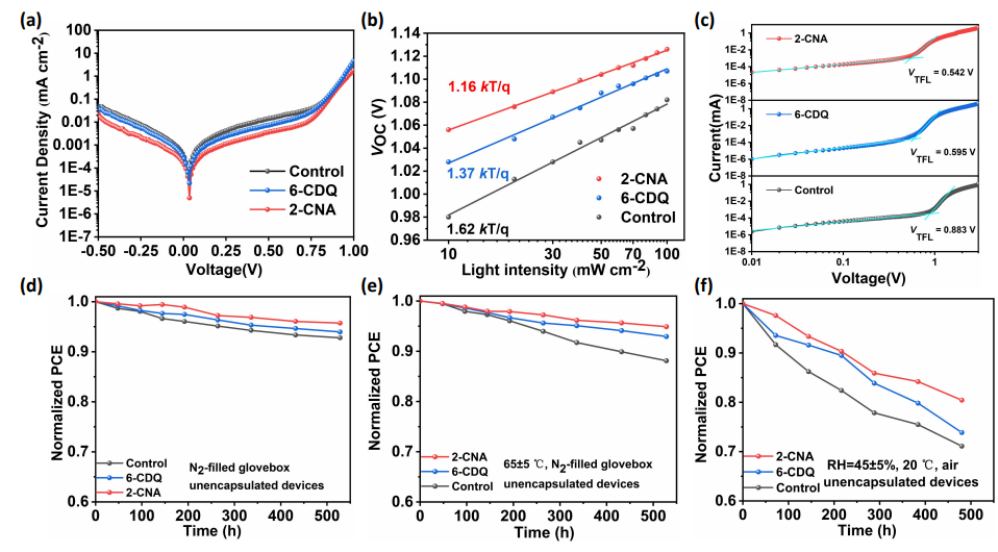}
    \caption{Defect suppression and \textcolor{black}{stability under storage, thermal, and ambient-humidity conditions} for control, 6-CDQ-, and 2-CNA-treated devices. 
    (a) Dark \textit{J}-\textit{V} characteristics, (b) light-intensity-dependent $V_{\mathrm{OC}}$ analysis, and 
    (c) SCLC analysis of electron-only devices. PCE retention of unencapsulated devices during storage 
    in a nitrogen-filled glovebox at (d) 25~$^{\circ}$C for 528~h and (e) 65~$^{\circ}$C for 528~h, 
    and (f) ambient-air exposure at 20~$^{\circ}$C and 45~$\pm$~5\% RH for 480~h.}
    \label{fig:fig-7}
\end{figure}

The device-level consequences of these differences were further supported by electrical and stability measurements. As shown in Figure~\ref{fig:fig-7}a-c, compared with the control, the 6-CDQ- and 2-CNA-treated devices exhibited lower dark current, reduced light-intensity-dependent \textit{V}$_\mathrm{OC}$ slopes~\cite{tress2018interpretation}, and lower trap-state densities~\cite{lecorre2021revealing}, with N$_t$ decreasing from 8.07 × $10^{15}$ $cm^{-3}$ for the control to 5.44 × $10^{15}$ and 4.96 × $10^{15}$ $cm^{-3}$ for 6-CDQ- and 2-CNA-treated devices, respectively. 
Stability tests under nitrogen storage, thermal stress, and ambient humidity further showed improved retention for the effective LEAP-selected additives~\cite{zhang2026photoswitchable,khenkin2020consensus}, as summarized in Figure~\ref{fig:fig-7}d-f. After storage in a nitrogen-filled glovebox at 25 °C for 528 h, the control, 6-CDQ-, and 2-CNA-treated devices retained 92.8\%, 94.0\%, and 95.7\% of their initial PCE, respectively. Under \textcolor{black}{thermal storage} at 65 °C in nitrogen for 528 h, the corresponding PCE retentions were 88.1\%, 92.9\%, and 94.9\%. After exposure to ambient air at 20 °C and 45 ± 5\% relative humidity~(RH) for 480 h, the devices retained 71.1\%, 73.8\%, and 80.4\% of their initial PCE, respectively. Together with the reduced dark current, lower light-intensity-dependent \textit{V}$_\mathrm{OC}$ slopes, and decreased trap-state densities, these results suggest that 2-CNA is associated with the most favorable \textcolor{black}{defect-suppression and stability characteristics under the tested storage, thermal, and ambient-humidity conditions} among the tested additives.

\section*{Discussion}\label{sec-discussion}
LEAP is best understood as a closed-loop discovery framework rather than as a stand-alone language-model application. In this workflow, literature-grounded mechanistic reasoning serves as an intermediate layer that is translated into interpretable molecular representation and coupled to uncertainty-aware candidate prioritization. This design is particularly useful for perovskite precursor additive discovery, where candidate quality depends not only on molecular structure, but also on mechanism-level effects on precursor coordination, crystallization, defect passivation, and environmental tolerance. By linking domain-specialized reasoning, hybrid molecular descriptors, Bayesian optimization, and experimental feedback, LEAP provides a practical route for converting literature-scale knowledge into experimentally actionable hypotheses. \textcolor{black}{Accordingly, the present validation is best interpreted as application-level repurposing of known small molecules, since their use as perovskite precursor additives was not identified in our literature-derived additive corpus or in exact-name/CAS literature searches.}

\textcolor{black}{We acknowledge that the absolute efficiencies of the approximately 1.55 eV devices remain below state-of-the-art values for this bandgap. The present experiments were designed to evaluate additive-induced changes within a fixed fabrication platform rather than to establish an efficiency record. Accordingly, additive effectiveness is assessed from controlled comparisons against devices fabricated under the same architecture, batch conditions, and measurement procedures, together with the observed batch-level performance trends and complementary film and device characterization.}

\textcolor{black}{The experimental characterization results also provide a post-validation link back to the initial Perovskite-RL mechanistic descriptors. In the representative descriptor rerun summarized in Table~S2, 6-CDQ was assigned high binding, hydrophobic-protection, and ion-interaction scores, whereas 2-CNA was dominated by high binding and ion-interaction scores with a higher predicted-effect score. These model-derived hypotheses are broadly consistent with the subsequent characterization: FTIR measurements showed interactions between both 6-CDQ and 2-CNA and PbI$_2$, with additional evidence of 2-CNA interaction with FAI; XRD, UV-Vis, and SEM measurements indicated improved crystallinity, absorption, and grain morphology; and dark \textit{J}–\textit{V}, light-intensity-dependent \textit{V}$_\mathrm{OC}$, and SCLC measurements showed reduced recombination signatures and lower trap-state densities for the effective additives. Therefore, although the current experiments do not constitute atomistic proof of every model-assigned descriptor, they close a practical prediction-verification-explanation loop by connecting LLM-derived coordination/ion-interaction hypotheses with experimentally observed precursor interactions, film-quality changes, and defect suppression. This distinction is important because LLM-derived mechanistic rationales can be incomplete or overconfident if used without constraints; in LEAP, these rationales are therefore treated as hypothesis-generating descriptors that must be filtered by expert feasibility review and evaluated against experimental characterization rather than as standalone mechanistic evidence.}

\textcolor{black}{The present study also defines the current scope and limitations of the framework. First, the prospective experimental validation was limited to three LEAP-prioritized additives, with one experimentally tested additive per round, in one device architecture, although 24 devices were fabricated for each tested condition. Therefore, the validation should be interpreted as a proof-of-concept demonstration of an expert-in-the-loop prioritization workflow rather than a statistically powered proof that LEAP can continuously identify high-quality additives across arbitrary candidate pools. Larger-scale prospective campaigns with multiple experimentally tested candidates per round will be required to quantify discovery efficiency, robustness, and failure modes. Importantly, a failed validation round would not be discarded in the LEAP workflow. Instead, the measured negative outcome would be appended to the training set as feedback, allowing the GP surrogate, acquisition scores, and subsequent candidate ranking to be updated in the next iteration. Thus, the present results should be viewed as an initial closed-loop case study, while rigorous assessment of robustness to failed or weak-performing rounds will require larger prospective validation campaigns.}
Second, LEAP currently operates in an expert-in-the-loop mode, and candidate selection still depends on expert review for chemical plausibility, commercial availability, precursor compatibility, safety, cost, and device-fabrication feasibility. 
\textcolor{black}{Because these expert decisions were recorded as qualitative feasibility judgments rather than as a structured accept/reject dataset for every model-prioritized candidate, the present study does not quantify a precise expert-rejection fraction or assign molecule-specific rejection reasons to all untested candidates. Future prospective LEAP campaigns should therefore maintain structured expert-review records using predefined categories so that the respective contributions of model ranking and expert feasibility assessment can be quantified more directly.}
Third, the mechanism-aware representation used here is tailored to perovskite additive engineering and has not yet been shown to generalize to other perovskite compositions, device structures, or materials-discovery domains. In addition, although the retrospective ablations provide diagnostic support for the usefulness of mechanism-aware descriptors, domain-adapted reasoning, and EI-based acquisition, these analyses were performed on a limited experimental dataset and should not be viewed as exhaustive proof of optimal representation or acquisition design. Further experimental-scale comparisons will be needed to quantify how these components affect discovery efficiency across broader candidate spaces. \textcolor{black}{Because the prospective validation combined model-guided ranking with expert feasibility review and changing candidate pools, the present experiments do not by themselves isolate the independent contribution of each decision component.}

More broadly, LEAP suggests a potentially transferable workflow for data-limited scientific problems in which prior knowledge is distributed across the literature and candidate validation remains experimentally costly. With appropriate domain adaptation and further validation, similar literature-grounded, expert-in-the-loop optimization workflows could be explored in other molecular and materials discovery tasks.

\section*{Conclusion}\label{sec-conclusion}
In summary, we developed LEAP as an expert-in-the-loop proof-of-concept workflow for perovskite precursor additive prioritization. The workflow integrates literature-grounded mechanistic reasoning, interpretable molecular representation, Bayesian optimization, expert feasibility review, and experimental feedback. By converting domain knowledge from the additive literature into mechanism-aware descriptors and combining them with uncertainty-aware candidate prioritization, LEAP provides a practical route for experimentally actionable molecular selection under low-data conditions.

In this case study, three rounds of validation identified later-round additives that improved device performance relative to the control, while materials and device characterizations were consistent with more favorable precursor interactions, film formation, defect suppression, and \textcolor{black}{stability under the tested storage, thermal, and ambient-humidity conditions}. These results provide preliminary evidence that literature-grounded reasoning can support additive prioritization when coupled with active learning and expert review. Further studies across larger candidate spaces, additional perovskite compositions, and different device architectures will be needed to establish the broader generality of this workflow.

\section*{Methods}
\subsection*{Data curation, model training, and \textcolor{black}{mechanism-consistency evaluation}}
To construct the domain-specific reasoning module used in LEAP, we curated a training corpus from 1,264 literature articles on perovskite precursor additives together with a candidate-molecule library assembled by domain experts. \textcolor{black}{Here, the candidate-molecule library refers to the auxiliary molecular-property component used for model adaptation and descriptor construction, whereas the prospective screening pools used in the LEAP validation rounds were generated separately by Perovskite-RL during the active-learning workflow.} From these sources, we extracted information relevant to additive identity, mechanism of action, reaction pathways, stability-related effects, and molecular physicochemical properties, thereby forming two complementary data components: a literature-derived mechanistic dataset and a molecular property dataset. After data cleaning, validation-set construction, and oversampling of the mechanistic component to alleviate token imbalance, the final SFT dataset contained 90,749 entries, together with an independent validation set of 1,000 entries. In addition, a reinforcement-learning dataset of 5,800 entries was assembled from the same literature source base.

The literature-derived mechanistic dataset contained 29,554 structured entries, and the auxiliary molecular-property dataset contained 33,141 entries describing physicochemical properties and functional groups. Five hundred records from each component formed the 1,000-entry validation set. Because the mechanistic component contained fewer tokens, it was oversampled before integration into the 90,749-entry SFT training set. A separate 5,800-entry dataset was assembled for RL. The domain-specialized language model, Perovskite-RL, was obtained by adapting Qwen3-32B~\cite{yang2025qwen3} through SFT followed by RL~\cite{ouyang2022training}.

For SFT, low-rank adaptation~(LoRA)~\cite{hu2022lora} was applied to the query, key, value, output, gate, and up projection matrices with rank 32, alpha 64, and dropout 0.05. Training used a learning rate of $3\times10^{-5}$, a cosine scheduler, a warmup ratio of 0.03, a per-device batch size of 1, and 16 gradient-accumulation steps (effective batch size 16). Training stopped at approximately 1.76 epochs when training loss had largely converged and validation loss had plateaued. The SFT model was merged before the RL stage, which used Group Relative Policy Optimization~(GRPO)~\cite{shao2024deepseekmath} with LoRA rank 16, alpha 32, and dropout 0.05 on the query, key, value, and output projection matrices. The RL learning rate was $2\times10^{-5}$ with a cosine scheduler and a warmup ratio of 0.05. GRPO used eight generations per prompt, a maximum completion length of 5,200 tokens, a total sequence length of 8,192, a KL-penalty coefficient of 0.05, and gradient clipping at 0.5. The reward combined answer correctness (the highest-weighted term), format compliance, coverage of reference-related content, and reasoning quality. A heuristic logic-density term rewarded logical connectives and inferential structure; partial credit was assigned for reasonable intermediate reasoning even when the final answer was incorrect. Anti-repetition and mild length penalties were also applied.

To evaluate whether the resulting model could provide reliable mechanistic assessments for downstream use in LEAP, we further constructed a \textcolor{black}{32-question mechanism-consistency evaluation set} from \textcolor{black}{11} perovskite-additive studies that were strictly excluded from the training corpus. From these unseen papers, we curated 32 multiple-choice questions to test whether the model could produce judgments consistent with the mechanistic interpretations reported in the original literature. \textcolor{black}{The questions and answer keys were finalized before model evaluation, and no question or answer key was modified based on model performance. The source papers were not used for SFT, RL, retrieval-augmented generation, or web-augmented evaluation.} The performance of the domain-specialized model was compared against its backbone model and several general-purpose baselines, and pairwise statistical comparisons were performed using exact McNemar tests followed by Holm-Bonferroni correction across the five baseline comparisons. \textcolor{black}{In addition, three human experts with perovskite research experience who were not involved in question generation or answer-key assignment independently answered the same questions under a closed-book setting.} This \textcolor{black}{evaluation} was used as a focused sanity check of the literature-grounded reasoning capability of the knowledge module before its deployment in downstream molecular representation and candidate prioritization.

\subsection*{LEAP workflow for molecular representation and candidate prioritization}
\textcolor{black}{In the prospective validation rounds, candidate molecules were generated by Perovskite-RL using mechanism-oriented additive-design prompts, rather than enumerated from a fixed commercial-catalog database. Generated molecules were standardized by name, CAS number, and SMILES where available; chemically implausible duplicates and candidates incompatible with precursor-solution processing were removed before model-based prioritization. The remaining molecules were passed into the LEAP representation and prioritization pipeline, which was designed for additive discovery rather than structural description alone.}
\textcolor{black}{To the best of our knowledge, none of the 150 candidates in the three released round-specific top-50 shortlists had previously been reported as perovskite precursor additives (0/150, 0\%).}
Instead of relying solely on conventional molecular descriptors or high-dimensional latent embeddings from the language model, LEAP derives explicit soft descriptors that capture five mechanism-relevant dimensions of additive function: binding mode, interfacial shielding, hydrophobic protection, ion-interaction potential, and electronic modulation. These descriptors are obtained through repeated literature-grounded reasoning and aggregation, allowing the representation to capture both dominant mechanistic tendencies and inference variability across candidate molecules.

For each molecule, Perovskite-RL generated 10 sampled reasoning trajectories at temperature 0.7. Outputs were mapped to predefined binary or categorical judgments for the five mechanism dimensions and an aggregated predicted-effect score. The means and standard deviations across the repeated outputs formed the soft descriptors, capturing both the inferred tendency and inference variability. The five mechanism scores alone defined the ``Mechanism soft'' representation in the retrospective ablation, whereas ``Full soft'' additionally included the predicted-effect score. Hard descriptors were calculated with RDKit and project-defined SMARTS/fragment rules, including molecular weight, LogP, topological polar surface area, hydrogen-bond donor and acceptor counts, rotatable-bond count, and counts of predefined functional-group motifs. Hard and soft descriptors were concatenated to form the hybrid representation. Constant features were removed, and the remaining features were standardized by the training-set mean and standard deviation in each round.

The active-learning workflow was initialized with 36 additives previously characterized in our laboratory. The Gaussian-process~(GP) surrogate used a constant-scaled Mat\'ern kernel with $\nu=1.5$ and a learnable WhiteKernel noise term. Hyperparameters were optimized by maximizing the log marginal likelihood with five optimizer restarts. The target was relative PCE change, $\Delta_{\mathrm{rel}}=(\mathrm{PCE}_{\mathrm{additive}}-\mathrm{PCE}_{\mathrm{control}})/\mathrm{PCE}_{\mathrm{control}}$. For each untested candidate, the GP returned a predicted mean response and predictive standard deviation. Expected improvement~(EI) used the 80th percentile of the relative PCE changes in the current training set as its current-best reference, with exploration offset $\xi=0.05$.

\textcolor{black}{Specifically, the GP surrogate takes the hybrid LEAP representation, consisting of deterministic molecular descriptors and Perovskite-RL-derived soft descriptors, as input and learns the relationship to the experimentally measured relative PCE change. For each untested candidate, the GP posterior provides both a predicted mean response and a predictive standard deviation, which are then combined in the expected-improvement acquisition function. After each validation round, the newly measured additive performance is appended to the hot-start dataset, the feature scaling and GP surrogate are updated, and the candidate pool is rescored for the next round.}

Candidate prioritization in LEAP is guided by the expected improvement acquisition function, which balances exploitation of candidates predicted to perform well against exploration of under-sampled regions of the candidate space. 
\textcolor{black}{At the current stage, the workflow operates in an expert-in-the-loop mode, in which model-prioritized candidates are further reviewed for chemical plausibility, commercial availability, precursor-solution compatibility, safety, cost, and device-fabrication feasibility before device validation.}
\textcolor{black}{In this proof-of-concept implementation, expert review was performed as a qualitative feasibility assessment rather than as a separately parameterized scoring model. The review considered chemical plausibility, commercial availability, precursor-solution compatibility, expected solubility and processability, safety, cost, and compatibility with the established device-fabrication workflow.}
Newly acquired experimental results are then incorporated into the next round of surrogate modelling and candidate selection, forming an expert-in-the-loop feedback process for iterative additive prioritization. For comparison, a non-LEAP selection route was also evaluated under the same device architecture and testing conditions, as described in the Results section.

Retrospective computational ablations were performed to assess the contributions of molecular representation, reasoning source, and decision policy. Unless otherwise specified, Gaussian-process surrogate models were trained using the 36 hot-start additives with relative PCE change as the target. Representation and reasoning-source ablations were evaluated by leave-one-out cross-validation using ranking- and error-based metrics, including Spearman correlation, top-k overlap, and RMSE~\cite{spearman1904proof}. The decision-policy ablation was performed over a generated candidate pool using the same trained surrogate, and compared candidate shortlists obtained by EI, predicted mean, predictive uncertainty, and random selection. These ablations were intended to analyze model and policy behavior retrospectively and did not constitute experimental validation of all shortlisted candidates.

For the reasoning-source ablation, soft descriptors were generated separately by Perovskite-RL and its Qwen3-32B backbone using the same descriptor schema, GP model, leave-one-out protocol, and target. Repeated outputs were summarized by their means and standard deviations before combination with the same hard descriptors. The hard-only baseline in this analysis was an internal reference for the reasoning-source ablation setting and was not intended to be numerically identical to the hard-feature baseline in the main-text representation ablation. For the decision-policy ablation, the GP trained on the 36 hot-start additives assigned predicted mean, predictive uncertainty, and EI values to the same generated candidate pool; top-50 lists were constructed separately by ranking each quantity or by random selection. These fixed-pool comparisons assessed prioritization behavior rather than prospective device performance. In the non-LEAP comparison route, three additives were selected by domain experts from the same candidate library using chemical intuition and practical feasibility, without model-guided active-learning ranking. The same commercial-availability, precursor-compatibility, device-fabrication, and testing constraints were applied. This three-candidate route was a practical reference, not a randomized or exhaustive benchmark.

In each validation round, molecules were ranked by EI and the top 50 formed a shortlist for expert feasibility review. The scored pools contained 25,639, 13,380, and 20,630 candidates in rounds 1--3, respectively. In round 3, functional-group prioritization reduced the final selection pool to 5,187 candidates before shortlist construction. Boc-DCPy, 6-CDQ, and 2-CNA occupied positions 25, 1, and 10 in their round-specific top-50 shortlists. The tested molecules were advanced after qualitative expert review rather than automatic selection of the top-ranked EI candidate. Expert decisions were not recorded as molecule-specific accept/reject labels for every untested candidate. After each round, measured relative PCE changes were added to the hot-start data, the GP was retrained, and the Perovskite-RL prompting template was updated with the new experimental observations. Soft descriptors were regenerated under the revised prompting context before candidates were rescored for the next round. Candidate identities were checked by name, CAS number, and SMILES where available. An exact-identity audit of the tested additives against the literature-derived additive corpus and prior perovskite-additive reports was used to distinguish application-level repurposing from rediscovery of previously reported precursor additives.

For a bounded descriptor reproducibility check, a representative panel in each round comprised the top 50 shortlist molecules, 50 randomly sampled candidates from the corresponding selection pool, and the experimentally tested additive. Each molecule was evaluated 10 times with the same descriptor schema, giving 302 molecule-round entries and 3,020 model responses; all responses were parsed into the predefined fields. The released round-specific shortlist dataset contains 50 entries per round with molecular identifiers, shortlist ranks, GP-predicted means, predictive uncertainties, EI values, mechanism scores, and experimental-advancement status. An entry not advanced in this campaign was not necessarily individually reviewed and rejected by experts.

\subsection*{Materials, device fabrication, and characterization}
Patterned indium tin oxide~(ITO) glass, PbI$_2$ (99.999\%), and NiO$_x$ nanoparticles (99.5\%) were purchased from Advanced Election Technology Co., Ltd. Bathocuproine~(BCP, 99\%), C$_{60}$ (99\%), MACl (99.5\%), CsI (99.99\%), MAI (99.5\%), and FAI (99.5\%) were purchased from Xi'an Yuri Solar Co., Ltd. (4-(7H-dibenzo[c,g]carbazol-7-yl)butyl)phosphonic acid~(4PADCB, $>$98\%) was purchased from TCI Corporation. N-Boc-4,6-dichloropyridin-2-amine~(Boc-DCPy, 97\%), 6-chloro-1,2,3,4-tetrahydroquinolin-2-one~(6-CDQ, 97\%), and 2-cyanoacetamide~(2-CNA, 98\%) were purchased from Jiangsu Aikon Biopharmaceutical R\&D Co., Ltd. Chlorobenzene (99.8\%), isopropanol (99.8\%), DMF (99.8\%), and DMSO (99.8\%) were purchased from Beijing J\&K Technology Co., Ltd. All reagents were used as received. The 1.67~M Cs$_{0.05}$MA$_{0.1}$FA$_{0.85}$PbI$_3$ precursor was prepared by dissolving 16.63~mg MACl, 21.63~mg CsI, 26.5~mg MAI, 243.63~mg FAI, and 807.3~mg PbI$_2$ in 1~mL DMF/DMSO (4:1, v/v). Each tested additive was introduced separately at 0.3~mg~mL$^{-1}$.

Devices had the structure ITO/\allowbreak NiO$_x$/\allowbreak 4PADCB/\allowbreak perovskite/\allowbreak C$_{60}$/\allowbreak BCP/\allowbreak Ag. Patterned ITO substrates were ultrasonicated sequentially in detergent, deionized water, acetone, and isopropanol for 30~min each, dried at 65~$^\circ$C overnight, and treated with UV--ozone for 5~min. An aqueous NiO$_x$ dispersion (10~mg~mL$^{-1}$) was spin-coated at 3000~rpm for 30~s and annealed in air at 150~$^\circ$C for 30~min. The substrates were transferred into a nitrogen glovebox, where 4PADCB (0.5~mg~mL$^{-1}$ in anhydrous ethanol) was spin-coated at 3000~rpm for 30~s and annealed at 100~$^\circ$C for 10~min. The perovskite precursor was spin-coated at 1000~rpm for 10~s (200~rpm~s$^{-1}$ acceleration) and then 5000~rpm for 35~s (1000~rpm~s$^{-1}$ acceleration). During the second step, 150~$\mu$L chlorobenzene was dropped 15~s before the end of spinning. Films were annealed at 120~$^\circ$C for 10~min. Finally, C$_{60}$ (26~nm), BCP (6~nm), and Ag (100~nm) were deposited by thermal evaporation. Electron-only devices for trap-density measurements had the structure ITO/SnO$_2$/perovskite/C$_{60}$/BCP/Ag.

FTIR spectra of 6-CDQ, 2-CNA, FAI, 6-CDQ+FAI, 2-CNA+FAI, 6-CDQ+PbI$_2$, and 2-CNA+PbI$_2$ were acquired with a Bruker Vertex 70 spectrometer. X-ray diffraction was measured with a Shimadzu XRD-7000 diffractometer, and top-view scanning electron microscopy images were obtained with a Hitachi SU8010 microscope; grain-size distributions were quantified from 50--100 grains per condition. Steady-state photoluminescence was measured on glass/perovskite films, while transient photocurrent and transient photovoltage were measured on complete devices. Direct optical bandgaps were estimated from UV--Vis absorption data by extrapolating the linear region of $(\alpha h\nu)^2$ versus $h\nu$ to the energy axis. Dark \textit{J}--\textit{V} and space-charge-limited-current measurements used the Keithley 2400 under dark conditions; light-intensity-dependent \textit{V}$_\mathrm{OC}$ was used to assess recombination.

Photovoltaic \textit{J}--\textit{V} curves of unencapsulated devices were measured with a Keithley 2400 SourceMeter under simulated AM 1.5G illumination (100~mW~cm$^{-2}$; CROWNTECH) at room temperature in ambient air. Reverse scans ran from 1.15 to $-0.10$~V and forward scans over the same window in the opposite direction, with 10~mV steps and 10~ms delay per step (nominal scan rate 1.0~V~s$^{-1}$). The active area was 0.1~cm$^2$, defined by a metal aperture mask; illumination was calibrated with a standard silicon reference cell. Both scan directions were recorded, but no quantitative hysteresis index was used. EQE spectra were acquired with an EnliTech QE-R instrument and integrated current densities were compared with \textit{J}--\textit{V}-derived \textit{J}$_\mathrm{SC}$. The champion 2-CNA device was held at 0.84~V for 60~s for a stabilized power-output measurement. Stability of unencapsulated devices was evaluated over 528~h in nitrogen at 25~$^\circ$C, 528~h in nitrogen at 65~$^\circ$C, and 480~h in ambient air at 20~$^\circ$C and 45~$\pm$~5\% relative humidity. These storage, thermal, and ambient-exposure tests involved neither continuous illumination nor maximum-power-point tracking. The same fabrication and testing conditions were used for LEAP and non-LEAP device comparisons.

\subsection*{Data analysis and statistics}
Photovoltaic performance was evaluated using both champion-device metrics and population statistics. For each device condition, \textcolor{black}{24 device-level measurements were obtained from four independent fabrication batches, with six devices per condition in each batch}, unless otherwise specified, and the results are reported as mean $\pm$ standard deviation where applicable. Box plots were used to summarize the distributions of \textit{V}$_\mathrm{OC}$, \textit{J}$_\mathrm{SC}$, FF, and PCE, with boxes representing the 25th-75th percentiles and center lines indicating the medians.

For the mechanism-consistency evaluation, model accuracy was calculated over all 32 multiple-choice questions, and the 95\% confidence interval for Perovskite-RL accuracy was estimated using the Wilson method. Pairwise comparisons against baseline models used two-sided exact McNemar tests with Holm--Bonferroni adjustment across the five model comparisons. For the retrospective representation ablation, bootstrap resampling was performed at the additive level over the 36 hot-start additives. In each replicate, representation-specific predictions were resampled and Spearman rank correlation, top-20\% overlap, and RMSE improvement relative to the hard-feature baseline were recalculated; the 2.5th and 97.5th percentiles formed the 95\% confidence interval. Device statistics were calculated from raw device-level values and are reported as mean $\pm$ standard deviation. Differences in PCE relative to the control were evaluated with two-sided Welch's $t$-tests, with mean differences and 95\% confidence intervals reported in the Supplementary Information. Batch-to-batch reproducibility was assessed separately from the four batch-level mean PCE values; the pooled device-level tests do not represent 24 independent fabrication batches. Trap-state densities were calculated from the trap-filled-limit voltage of electron-only devices using the standard space-charge-limited-current formalism.

\section*{Data availability}

The data supporting the findings of this study are available within the article, the Supplementary Information, and the accompanying public data repository at \url{https://github.com/WD928/LEAP}. \textcolor{black}{The repository includes the benchmark, ablation, and released round-specific top-50 validation-shortlist identities, positions, GP/EI annotations, and mechanism-score annotations supporting the revised analyses.} The SFT and GRPO training datasets for Perovskite-RL are available at \url{https://huggingface.co/datasets/JH976/Perovskite-RL}. \textcolor{black}{Additional raw experimental data are stored on institutional servers and are available from the corresponding authors upon reasonable request.}

\section*{Code availability}

The Perovskite-RL model weights are available at \url{https://huggingface.co/JH976/Perovskite-RL}.

\bibliographystyle{unsrt}
\bibliography{references}

\vspace{36pt}
\noindent\textbf{Acknowledgement:}
The work is supported by the National Natural Science Foundation of China (No.62476278, No.11934020), Beijing Natural Science Foundation (No.Z250005, No.Z260002) and the National Key R\&D Program of China (Grants No. 2024YFA1408601). Computational resources have been provided by the Physical Laboratory of High Performance Computing at Renmin University of China.
\\

\noindent\textbf{Author contributions:}  
Z.F.G., C.M., and Z.Y.L. contributed to the ideation and design of the research; X.D.W., Z.R.C., and Z.F.G. performed the research; X.D.W., Z.R.C., P.J.G., Z.F.G., C.M., and Z.Y.L. wrote and edited the paper; all authors contributed to the research discussions. \\

\noindent\textbf{Corresponding authors:} 
Correspondence and requests for materials should be addressed to Ze-Feng Gao (zfgao@ruc.edu.cn), Cheng Mu (cmu@ruc.edu.cn), and  Zhong-Yi Lu (zlu@ruc.edu.cn). \\

\noindent\textbf{Competing interests:}
The authors declare no competing interests.\\

\noindent\textbf{Supplementary information:}
The supplementary information is attached.

\clearpage

\appendix
\clearpage          
\setcounter{page}{1} 
\renewcommand{\thepage}{S\arabic{page}} 

\setcounter{section}{0} 
\renewcommand{\thesection}{S\arabic{section}} 

\setcounter{table}{0}   
\renewcommand{\thetable}{S\arabic{table}}     

\setcounter{figure}{0}  
\renewcommand{\thefigure}{S\arabic{figure}}   

\begin{center}
    \vspace*{1cm} 
    
    {\Large \textbf{Supplementary Information for:}} \\[0.5em]
    
    {\Large \textbf{A closed-loop framework for perovskite precursor additive discovery}} \\[1em]
    
    \vspace{1cm} 
\end{center}

\section{Candidate-selection and descriptor data}
To improve transparency of the candidate-selection route, we summarized the round-specific candidate pools and validation-shortlist positions of the experimentally tested additives in Table~\ref{tab:tab-s1}. 
\textcolor{black}{Within this feasibility boundary, Boc-DCPy was advanced because its pyridyl/carbamate heteroatoms and chlorinated tert-butoxycarbonyl scaffold provided a combined binding and hydrophobic-protection hypothesis while remaining commercially and experimentally accessible. The Round-2 candidate 6-CDQ was advanced because it occupied the first shortlist position and combined a lactam-based coordination hypothesis with high hydrophobic-protection and ion-interaction descriptor scores. The Round-3 candidate 2-CNA was advanced from the prioritized pool because its compact cyanoamide structure provided multiple polar interaction sites, high binding and ion-interaction scores, and the highest predicted-effect score among the three tested additives, together with acceptable availability and precursor-solution compatibility. These selection rationales describe the hypotheses and feasibility considerations used to advance the molecules; they are not presented as independent proof of the proposed mechanisms. Because the expert review was conducted as a qualitative feasibility assessment during this proof-of-concept study, molecule-by-molecule accept/reject labels and molecule-specific rejection reasons were not recorded retrospectively for all higher-ranked candidates.}

\begin{table}[h]
\centering
\caption{Round-specific candidate pools and validation-shortlist positions for the LEAP-selected additives. The shortlist rank denotes the position of the experimentally tested molecule in the corresponding round-level top-50 validation shortlist.}
\label{tab:tab-s1}
\resizebox{\linewidth}{!}{
\begin{tabular}{lccclc}
\hline
Round & Scored & Selection pool & Selection rule & Additive & Shortlist rank \\
\hline
1 & 25,639 & 25,639 & All scored candidates & Boc-DCPy & 25 \\
2 & 13,380 & 13,380 & All scored candidates & 6-CDQ & 1 \\
3 & 20,630 & 5,187 & Prioritized candidates & 2-CNA & 10 \\
\hline
\end{tabular}
}
\end{table}

\textcolor{black}{To the best of our knowledge, none of the 150 candidates in the three released round-specific top-50 shortlists had previously been reported as perovskite precursor additives (0/150, 0\%). Boc-DCPy, 6-CDQ, and 2-cyanoacetamide (2-CNA, CAS 107-91-5) were present in the auxiliary molecular-property/mechanistic training component used for general molecular-mechanism learning, but they were not identified as previously reported perovskite precursor additives in the literature-derived 1,264-paper additive corpus. We therefore describe these validation examples as application-level repurposing of known molecules for previously unreported perovskite precursor-additive use, rather than discovery of new chemical entities.}

\textcolor{black}{For transparency, the released round-specific top-50 validation-shortlist identities and positions, together with the associated GP/EI and mechanism-score annotations, are provided in the public repository as the Supplementary Data file.}
\textcolor{black}{The file contains exactly 150 rows, with 50 released shortlist entries for each round, and reports molecule names, CAS numbers, SMILES strings, the actual round-specific \texttt{top50\_rank}, GP-predicted mean response, predictive uncertainty, expected-improvement score, Perovskite-RL mechanism scores, experimental-validation status, the tested-additive label, the general expert-review boundary, and experimental-advancement status. The three experimentally tested additives retain the manuscript shortlist positions of 25, 1, and 10 for Boc-DCPy, 6-CDQ, and 2-CNA, respectively. The numerical GP/EI fields are aligned to the original publicly released round-specific shortlist order recorded by \texttt{top50\_rank}. Because molecule-specific rejection notes for every untested candidate were not separately archived, \texttt{not\_advanced\_in\_this\_campaign} indicates only that an entry did not proceed to device validation and does not represent a formal molecule-specific expert rejection.}

The descriptor means for the experimentally tested additives are summarized in Table~\ref{tab:tab-s2}. 
\textcolor{black}{Boc-DCPy showed high hydrophobic-protection and ion-interaction scores, 6-CDQ showed high binding, hydrophobic-protection, and ion-interaction scores, and 2-CNA showed high binding and ion-interaction scores but low interfacial-shielding and hydrophobic-protection scores. As a qualitative example of the descriptor-mapping procedure, we examined the Perovskite-RL outputs for 2-CNA. In the representative rerun, the model identified the cyano group and amide carbonyl of 2-cyanoacetamide as plausible Lewis-base interaction sites for Pb-related precursor or defect sites, and assigned the amide NH2 group to possible hydrogen-bonding or ionic interactions with precursor ionic species. These qualitative assignments correspond to the high binding-mode and ion-interaction scores listed in Table~\ref{tab:tab-s2}. Conversely, the model assigned low interfacial-shielding and hydrophobic-protection scores because 2-CNA is a small polar molecule without a bulky hydrophobic skeleton, and a low electronic-modulation score because it lacks an extended conjugated framework. This example illustrates how the condensed model rationale was mapped onto the predefined soft-descriptor fields.}

\begin{table}[h]
\centering
\caption{Soft-descriptor means for the experimentally tested additives in the representative Perovskite-RL rerun. Each molecule was sampled 10 times, and all responses were successfully parsed.}
\label{tab:tab-s2}
\resizebox{\linewidth}{!}{
\begin{tabular}{lcccccccc}
\hline
Round & Additive & Parsed samples & Binding & Interfacial shielding & Hydrophobic protection & Ion interaction & Electronic modulation & Predicted effect \\
\hline
1 & Boc-DCPy & 10/10 & 0.80 & 0.40 & 1.00 & 0.90 & 0.00 & 0.40 \\
2 & 6-CDQ & 10/10 & 0.90 & 0.20 & 1.00 & 1.00 & 0.20 & 0.30 \\
3 & 2-CNA & 10/10 & 1.00 & 0.00 & 0.00 & 1.00 & 0.00 & 0.60 \\
\hline
\end{tabular}
}
\end{table}

\section{Retrospective representation ablation uncertainty}

The bootstrap results are summarized in Table~\ref{tab:tab-s3}. 
Consistent with the point estimates in Figure~3 of the main text, the hybrid LEAP representation showed the highest Spearman correlation point estimate and a small positive RMSE improvement relative to the hard-feature baseline. However, the confidence intervals were broad because the retrospective analysis contained only 36 additives. These results therefore support the use of the hybrid representation as a balanced and chemically interpretable feature space, but should not be interpreted as evidence of uniform superiority across all metrics.

\begin{table}[h]
\centering
\caption{Bootstrap confidence intervals for retrospective representation ablation. Spearman rank correlation, top-20\% overlap, and RMSE improvement were evaluated using the same 36-additive retrospective setting as in Figure~3 of the main text. RMSE improvement is defined as RMSE(hard-only) $-$ RMSE(current), so larger values indicate lower prediction error than the hard-feature baseline.}
\label{tab:tab-s3}
\resizebox{\linewidth}{!}{
\begin{tabular}{lccc}
\hline
Representation & Spearman & Top-20\% overlap & RMSE improvement \\
 & (95\% CI) & (95\% CI) & (95\% CI) \\
\hline
Hard & 0.274 (-0.042 to 0.541) & 0.125 (0.000 to 0.500) & 0.000 \\
Mechanism soft & 0.335 (0.027 to 0.590) & 0.500 (0.125 to 0.625) & -0.0131 (-0.0422 to 0.0106) \\
Full soft & 0.393 (0.073 to 0.648) & 0.375 (0.000 to 0.750) & -0.0080 (-0.0329 to 0.0120) \\
Hybrid (LEAP) & 0.394 (0.109 to 0.625) & 0.375 (0.000 to 0.625) & 0.0007 (-0.0134 to 0.0088) \\
\hline
\end{tabular}
}
\end{table}

\section{Retrospective reasoning-source ablation}
\begin{figure}[htbp]
    \centering
    \includegraphics[width=0.95\textwidth]{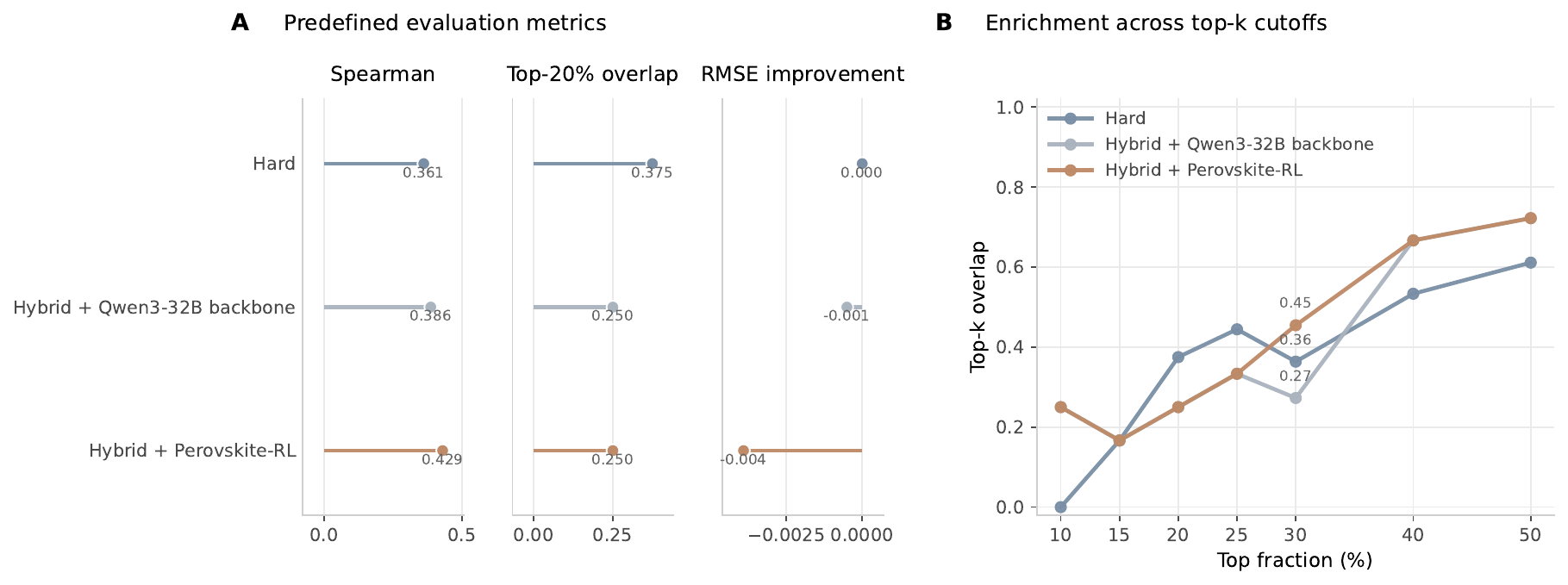}
    \caption{Reasoning-source ablation for LEAP soft descriptors. Perovskite-RL- and Qwen3-32B-derived descriptors were compared under the same hybrid Gaussian-process leave-one-out setting using 36 hot-start additives. (A) Spearman correlation, top-20\% overlap, and RMSE improvement. (B) Top-k overlap across enrichment cutoffs.}
    \label{fig:fig-s1}
\end{figure}

The reasoning-source comparison is shown in Figure~\ref{fig:fig-s1}, including the predefined evaluation metrics and top-$k$ overlap across enrichment cutoffs.

Under this controlled hybrid-representation setting, Perovskite-RL-derived descriptors showed a higher Spearman rank correlation than descriptors generated by the Qwen3-32B backbone, suggesting better global ranking consistency in this retrospective analysis. However, this difference did not translate into consistent gains in the strict top-20\% overlap or RMSE improvement relative to the hard-feature baseline. Because the 36-additive retrospective set makes strict top-k recovery sensitive to one-candidate changes, we further evaluated top-k overlap across a range of enrichment cutoffs. This sweep showed that the apparent enrichment advantage of Perovskite-RL was more visible at broader cutoffs, particularly around the top-30\% region, while both hybrid representations showed higher overlap than the hard-only representation at broader top-40\% and top-50\% cutoffs. These results suggest that domain-adapted reasoning may provide a modest and cutoff-dependent ranking or enrichment advantage, rather than a uniform improvement across all prioritization metrics.

\section{Retrospective decision-policy ablation}

\begin{figure}[htbp]
    \centering
    \includegraphics[width=0.98\textwidth]{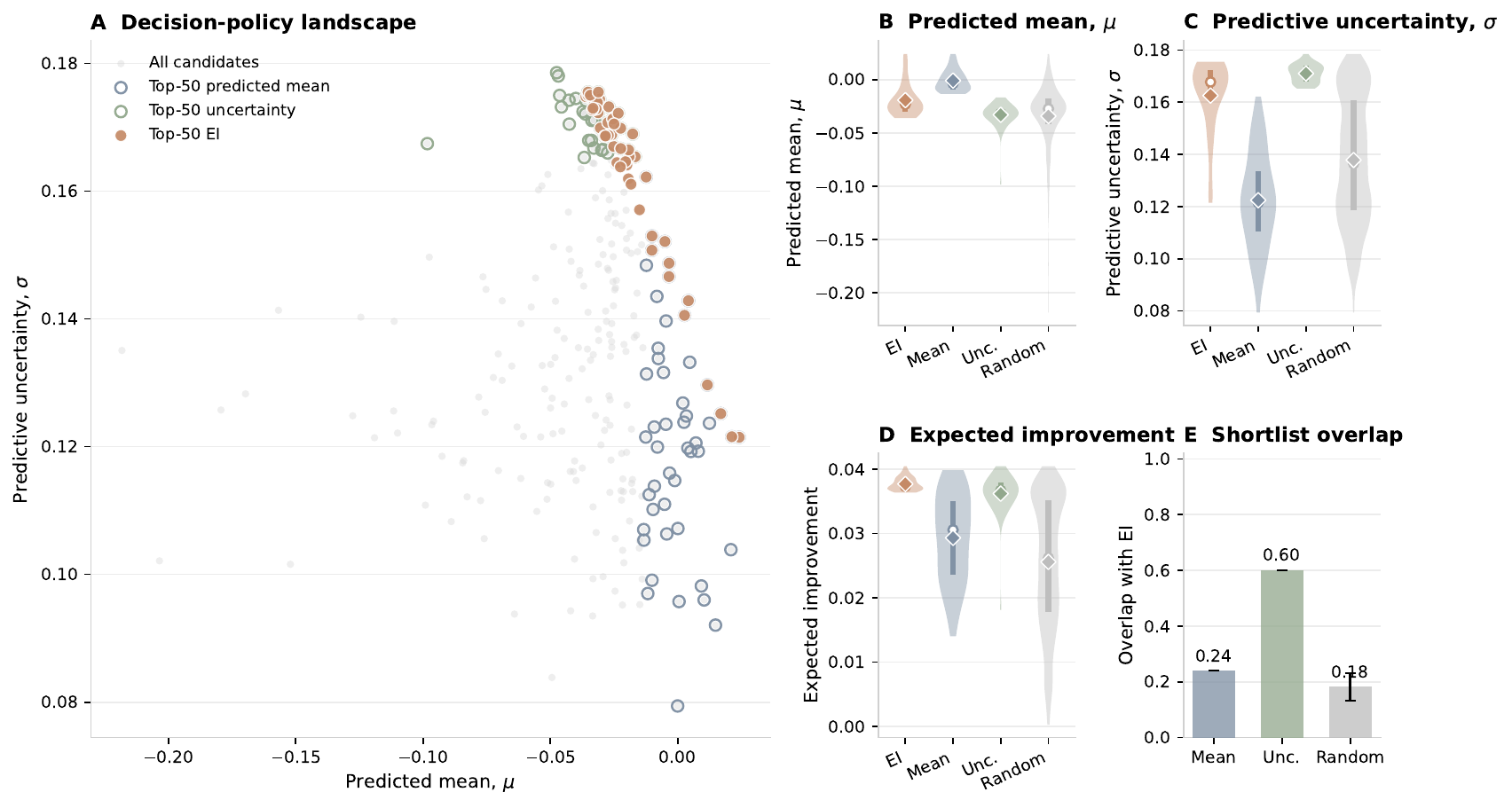}
    \caption{Retrospective decision-policy ablation for LEAP candidate prioritization. Using the same hybrid representation and Gaussian-process surrogate, candidates were ranked by expected improvement, predicted mean response, predictive uncertainty, or random selection. (A) Decision-policy landscape. (B–D) Distributions of predicted mean, uncertainty, and expected improvement for top-50 shortlists. (E) Overlap with the expected-improvement shortlist.}
    \label{fig:fig-s2}
\end{figure}

The decision-policy landscape, score distributions, and shortlist overlap are shown in Figure~\ref{fig:fig-s2}.
The mean-only policy selected candidates with the highest average predicted response, but with substantially lower predictive uncertainty. In contrast, the uncertainty-only policy selected candidates with the highest average uncertainty, but with lower predicted response. EI produced the highest average expected improvement while maintaining an intermediate predicted mean--uncertainty profile. Specifically, the EI-selected top-50 shortlist had a mean predicted response of $-0.019$, a mean predictive uncertainty of $0.163$, and a mean EI of $0.0377$. The mean-only shortlist had a higher mean predicted response ($-0.0009$) but lower uncertainty ($0.122$) and lower mean EI ($0.0293$), whereas the uncertainty-only shortlist had the highest uncertainty ($0.171$) but lower predicted response ($-0.033$). Random top-50 shortlists showed lower average EI ($0.0252 \pm 0.0014$), providing a background reference for the candidate pool.

The shortlist overlap analysis further showed that EI was not equivalent to either pure exploitation or pure exploration. The predicted-mean shortlist overlapped with the EI shortlist by 24\%, whereas the uncertainty-only shortlist showed 60\% overlap with the EI shortlist. Random shortlists showed an average overlap of 18\% with the EI shortlist. These results are consistent with EI incorporating exploration through predictive uncertainty while still retaining predicted-response information, supporting its use as a balanced exploitation-exploration decision policy in this LEAP implementation.

\section{Supplementary photovoltaic and defect-analysis data}
\begin{figure}[h]
    \centering
    \includegraphics[width=0.72\textwidth]{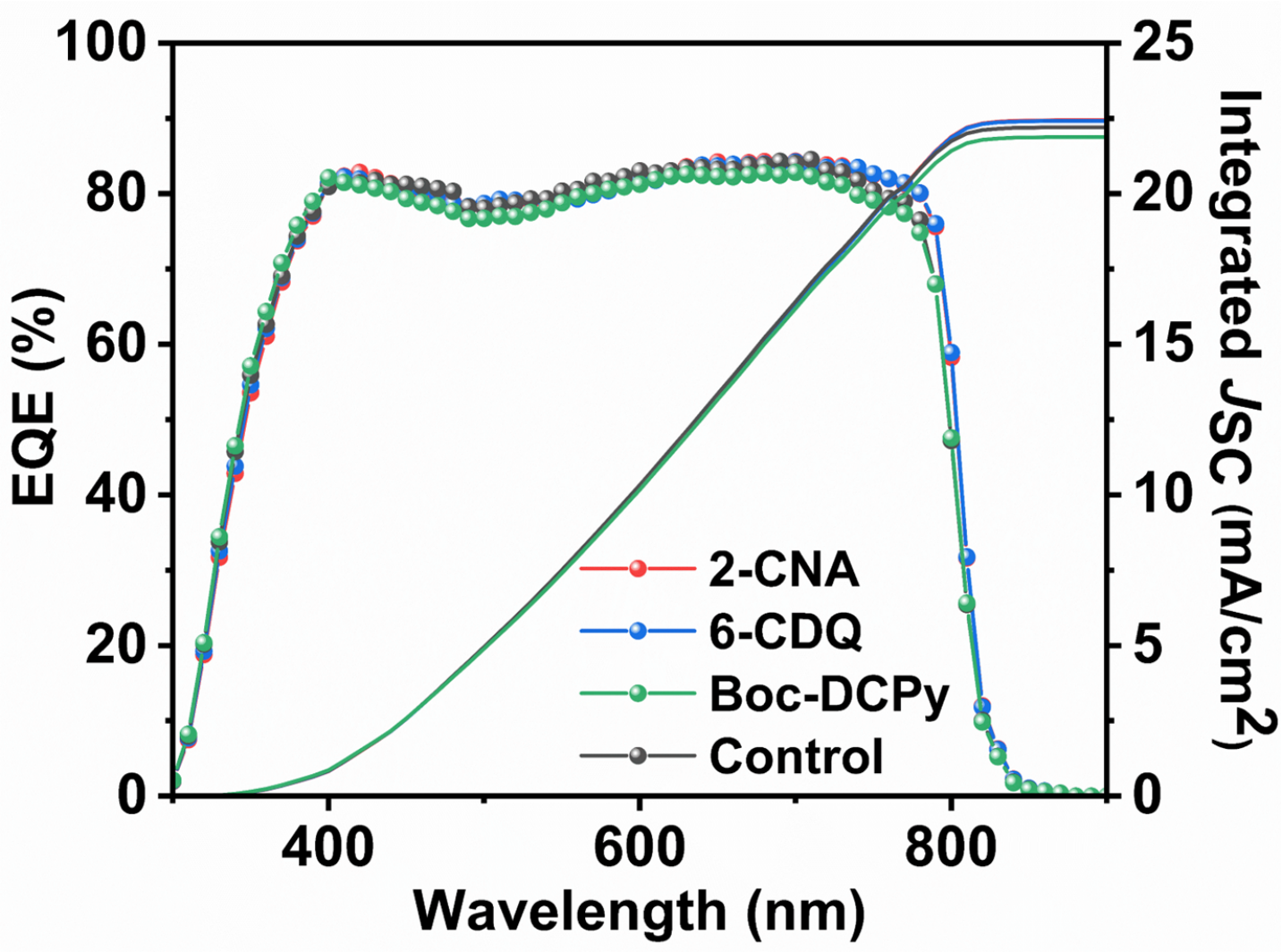}
    \caption{EQE spectra of control and additive-treated perovskite solar cells. Integrated current densities are consistent with the \textit{J}-\textit{V}-derived JSC values.}
    \label{fig:fig-s3}
\end{figure}

The EQE spectra of the control and additive-treated devices were measured to verify the photocurrent response obtained from the \textit{J}-\textit{V} measurements. As shown in Figure~\ref{fig:fig-s3}, 

the integrated current densities derived from the EQE spectra are consistent with the corresponding \textit{J}-\textit{V} results, supporting the reliability of the photovoltaic parameters reported in the main text. The photovoltaic parameters of the representative champion devices are summarized in Table~\ref{tab:tab-s4}. 
\begin{table}[h]
\centering
\caption{Photovoltaic parameters of representative champion PSCs with different additives.}
\label{tab:tab-s4}
\begin{tabular}{lcccc}
\hline
Sample & $V_{\mathrm{OC}}$ (V) & $J_{\mathrm{SC}}$ (mA cm$^{-2}$) & FF (\%) & PCE (\%) \\
\hline
Control & 1.090 & 23.77 & 76.57 & 19.85 \\
Boc-DCPy & 1.068 & 22.85 & 72.13 & 17.60 \\
6-CDQ & 1.101 & 23.87 & 78.24 & 20.57 \\
2-CNA & 1.128 & 23.92 & 79.00 & 21.32 \\
\hline
\end{tabular}
\end{table}
The statistical photovoltaic parameters extracted from \textcolor{black}{24 device-level measurements for each group, obtained from four independent fabrication batches with six devices per group in each batch,} are summarized in Table~\ref{tab:tab-s5}. 
\begin{table}[h]
\centering
\caption{Photovoltaic statistics of PSCs incorporating various additives ($n=24$ device-level measurements per group\textcolor{black}{, comprising six devices per group in each of four independent fabrication batches}). Values are reported as mean $\pm$ standard deviation. $\Delta$PCE, 95\% confidence intervals, and two-sided Welch's $t$-test $P$ values were calculated relative to the control group using raw device-level PCE values.}
\label{tab:tab-s5}
\resizebox{\textwidth}{!}{
\begin{tabular}{lccccccc}
\hline
Sample & $V_{\mathrm{OC}}$ (V) & $J_{\mathrm{SC}}$ (mA cm$^{-2}$) & FF (\%) & PCE (\%) & $\Delta$PCE vs control & 95\% CI for $\Delta$PCE & Welch $P$ value \\
\hline
Control & $1.082 \pm 0.007$ & $23.35 \pm 0.17$ & $74.30 \pm 1.08$ & $19.25 \pm 0.28$ & -- & -- & -- \\
Boc-DCPy & $1.053 \pm 0.008$ & $22.26 \pm 0.41$ & $69.06 \pm 2.38$ & $16.76 \pm 0.58$ & $-2.49$ & $-2.75$ to $-2.22$ & $2.23 \times 10^{-19}$ \\
6-CDQ & $1.100 \pm 0.006$ & $23.86 \pm 0.15$ & $77.05 \pm 0.60$ & $20.13 \pm 0.25$ & $+0.89$ & $0.73$ to $1.04$ & $4.44 \times 10^{-15}$ \\
2-CNA & $1.123 \pm 0.004$ & $23.84 \pm 0.16$ & $78.30 \pm 0.99$ & $20.87 \pm 0.25$ & $+1.62$ & $1.47$ to $1.77$ & $4.78 \times 10^{-25}$ \\
\hline
\end{tabular}
}
\end{table}
Trap-state densities were calculated from the trap-filled limit voltage obtained from the SCLC measurements, and the corresponding parameters are summarized in Table~\ref{tab:tab-s6}.
\begin{table}[!htbp]
\centering
\caption{Trap density ($N_t$) derived from electron-only SCLC measurements.}
\label{tab:tab-s6}
\begin{tabular}{lcccccc}
\hline
Sample & $\varepsilon_0$ ($10^{-12}$ F m$^{-1}$) & $\varepsilon$ & $V_{\mathrm{TFL}}$ (V) & $e$ ($10^{-19}$ C) & $L$ (nm) & $N_t$ ($10^{15}$ cm$^{-3}$) \\
\hline
Control & 8.85 & 46.5 & 0.883 & 1.6 & 750 & 8.07 \\
6-CDQ & 8.85 & 46.5 & 0.595 & 1.6 & 750 & 5.44 \\
2-CNA & 8.85 & 46.5 & 0.542 & 1.6 & 750 & 4.96 \\
\hline
\end{tabular}
\end{table}

\section{Details of the non-LEAP comparison route}
\begin{figure}[h]
    \centering
    \includegraphics[width=\textwidth]{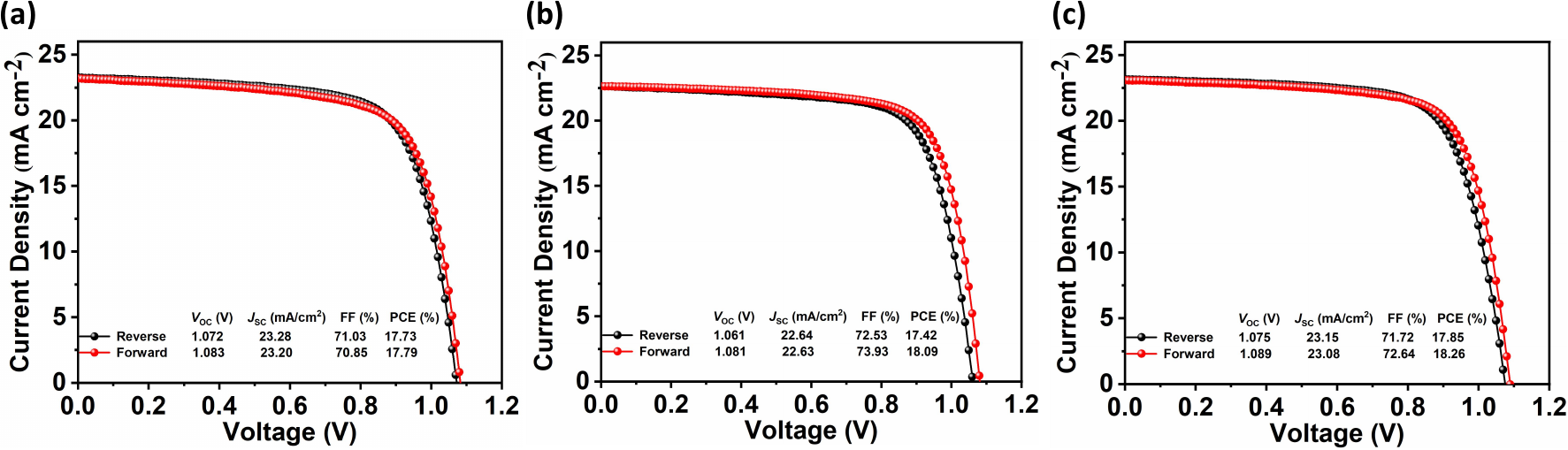}
    \caption{Representative \textit{J}-\textit{V} curves of non-LEAP-selected additives. 
    (a) 4-Hydroxy-8-(trifluoromethyl)quinoline, 
    (b) sodium \textit{m}-hydroxybenzenesulfonate, and 
    (c) 5-bromo-2-chloro-4-fluoroaniline.}
    \label{fig:fig-s4}
\end{figure}

Three representative additives were selected in the non-LEAP route, corresponding to three sequential validation rounds: 4-hydroxy-8-(trifluoromethyl)quinoline, sodium \textit{m}-hydroxybenzenesulfonate, and 5-bromo-2-chloro-4-fluoroaniline. The representative \textit{J}-\textit{V} curves of these non-LEAP-selected additives are shown in Figure~\ref{fig:fig-s4}. 

For 4-hydroxy-8-(trifluoromethyl)quinoline, the representative device exhibited a forward-scan \textit{V}$_\mathrm{OC}$ of 1.083 V, a \textit{J}$_\mathrm{SC}$ of 23.20 mA cm$^{-2}$, a fill factor of 70.85\%, and a PCE of 17.79\%. For sodium m-hydroxybenzenesulfonate, the representative device showed a forward-scan \textit{V}$_\mathrm{OC}$ of 1.081 V, a \textit{J}$_\mathrm{SC}$ of 22.63 mA cm$^{-2}$, a fill factor of 73.93\%, and a PCE of 18.09\%. For 5-bromo-2-chloro-4-fluoroaniline, the representative device yielded a forward-scan \textit{V}$_\mathrm{OC}$ of 1.089 V, a \textit{J}$_\mathrm{SC}$ of 23.08 mA cm$^{-2}$, a fill factor of 72.64\%, and a PCE of 18.26\%. These results show that, within this limited three-candidate reference route, the tested non-LEAP candidates did not reach the device performances obtained for the later-round LEAP-prioritized candidates.

Although this comparison route does not constitute an exhaustive \textcolor{black}{comparison} over the full candidate space, it provides a practical reference for interpreting the LEAP-guided validation results. In this limited comparison, the LEAP-guided route moved from an initially ineffective candidate to later-round additives with improved device performance, whereas the three tested non-LEAP candidates did not show a comparable enhancement trajectory. This contrast is consistent with a possible benefit of literature-grounded, feedback-driven prioritization, but it should not be interpreted as a definitive demonstration of superiority over expert-guided selection.

\section{\textcolor{black}{Batch-level device reproducibility}}
\textcolor{black}{Across the four independent fabrication batches, Boc-DCPy consistently reduced PCE relative to the control, whereas 6-CDQ and 2-CNA consistently increased PCE, as shown in Figure~\ref{fig:fig-s5}.}
\begin{center}
    \centering
    \includegraphics[width=\textwidth]{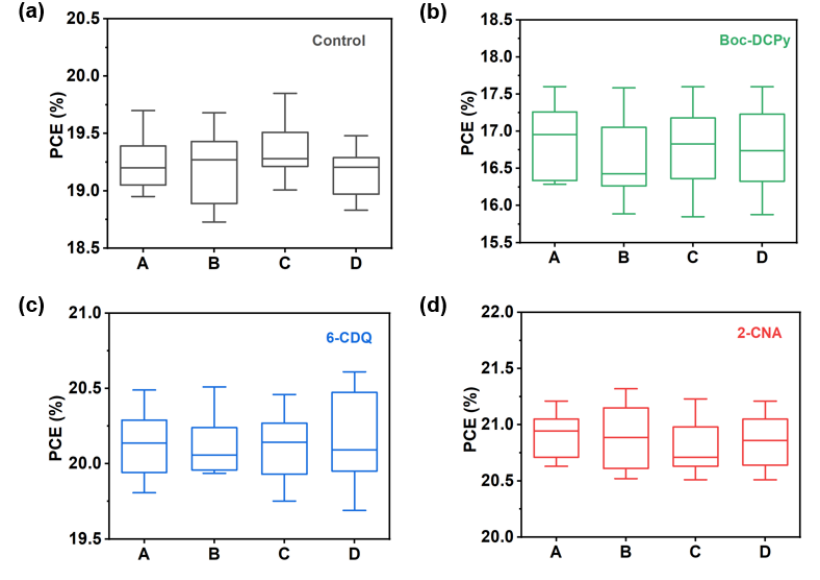}
    \captionsetup{labelfont={color=black},textfont={color=black}}
    \captionof{figure}{\textcolor{black}{Batch-level reproducibility of device PCE. PCE distributions are shown for the control, Boc-DCPy, 6-CDQ, and 2-CNA groups across four independent fabrication batches (A--D), with six devices per group in each batch.}}
    \label{fig:fig-s5}
\end{center}

\section{\textcolor{black}{Stabilized power output}}
\textcolor{black}{The champion 2-CNA device was held at 0.84 V for 60 s and yielded a stabilized PCE of 19.64\%, as shown in Figure~\ref{fig:fig-s6}.}
\begin{center}
    \centering
    \includegraphics[width=0.72\textwidth]{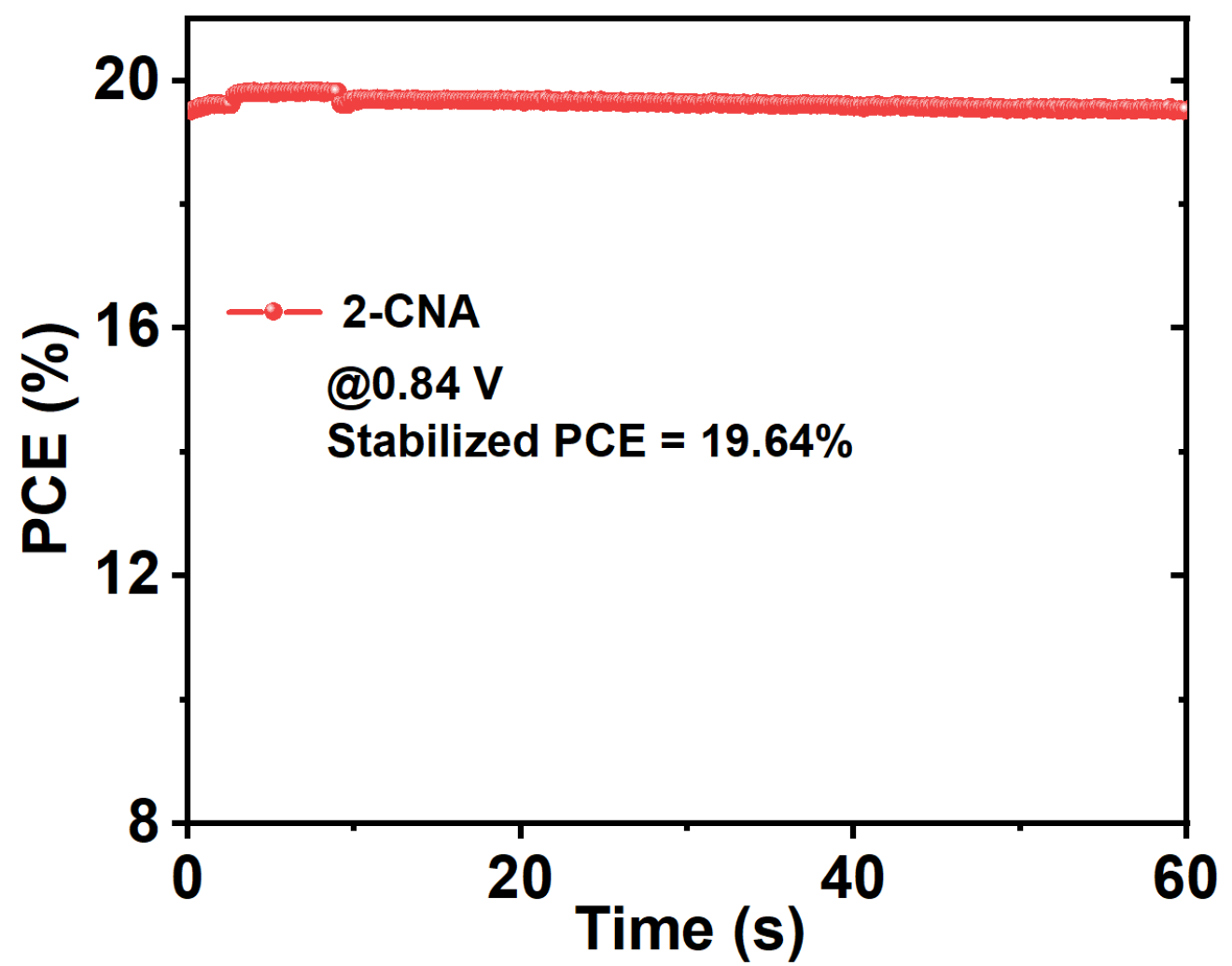}
    \captionsetup{labelfont={color=black},textfont={color=black},labelsep=colon}
    \captionof{figure}{\textcolor{black}{Stabilized power output of the champion 2-CNA device measured at 0.84 V for 60 s, yielding a stabilized PCE of 19.64\%.}}
    \label{fig:fig-s6}
\end{center}

\section{\textcolor{black}{Photophysical and grain-size analyses}}
\textcolor{black}{The PL intensity increased from the control to 6-CDQ and 2-CNA films. The TPC decay times were 0.882, 0.691, and 0.596 $\mu$s, while the TPV lifetimes were 10.3, 10.5, and 11.7 $\mu$s for the control, 6-CDQ, and 2-CNA samples, respectively, as shown in Figure~\ref{fig:fig-s7}.}
\begin{center}
    \centering
    \includegraphics[width=0.98\textwidth]{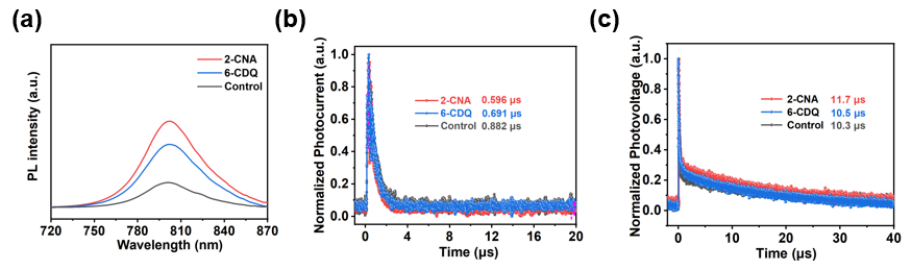}
    \captionsetup{labelfont={color=black},textfont={color=black},labelsep=colon}
    \captionof{figure}{\textcolor{black}{Photophysical characterization of control, 6-CDQ-, and 2-CNA-treated samples. (a) Steady-state PL spectra, (b) TPC decay traces, and (c) TPV decay traces.}}
    \label{fig:fig-s7}
\end{center}

\textcolor{black}{SEM-based grain-size analysis gave mean grain sizes of $0.69 \pm 0.21$ $\mu$m for the control, $0.77 \pm 0.30$ $\mu$m for 6-CDQ, and $0.79 \pm 0.26$ $\mu$m for 2-CNA (Figure~\ref{fig:fig-s8}).}
\begin{center}
    \centering
    \includegraphics[width=0.98\textwidth]{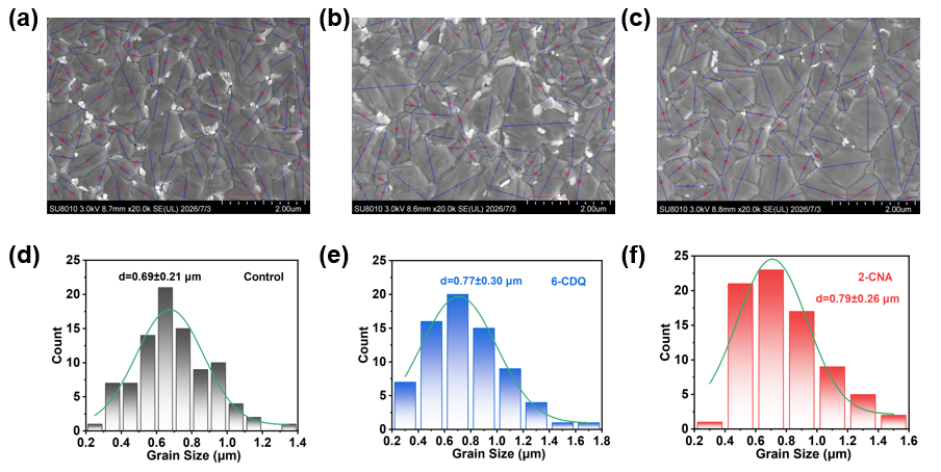}
    \captionsetup{labelfont={color=black},textfont={color=black},labelsep=colon}
    \captionof{figure}{\textcolor{black}{Top-view SEM images and quantitative grain-size analysis of perovskite films. (a--c) SEM images and (d--f) corresponding grain-size distributions for the control, 6-CDQ, and 2-CNA films, respectively.}}
    \label{fig:fig-s8}
\end{center}

\textcolor{black}{Tauc analysis yielded direct optical bandgaps of 1.543 eV for the control film, 1.543 eV for the 6-CDQ-treated film, and 1.545 eV for the 2-CNA-treated film (Figure~\ref{fig:fig-s9}), confirming that the investigated composition is an approximately 1.55 eV perovskite system.}
\begin{center}
    \centering
    \includegraphics[width=0.98\textwidth]{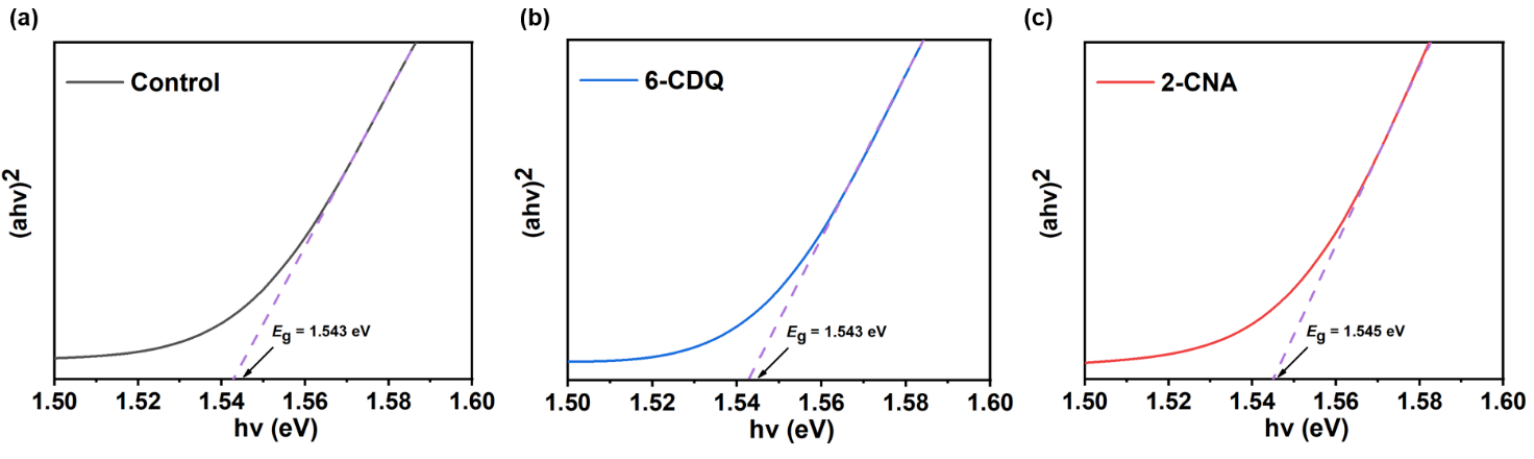}
    \captionsetup{labelfont={color=black},textfont={color=black},labelsep=colon}
    \captionof{figure}{\textcolor{black}{Tauc plots derived from UV--Vis absorption spectra of (a) control, (b) 6-CDQ-treated, and (c) 2-CNA-treated perovskite films. The extrapolated direct optical bandgaps are 1.543, 1.543, and 1.545 eV, respectively.}}
    \label{fig:fig-s9}
\end{center}

\end{document}